\documentclass[sn-mathphys-num, iicol]{sn-jnl}

\usepackage{adjustbox} 
\usepackage{subcaption} 
\usepackage{tikz} 
\usepackage[edges]{forest} 
\definecolor{hidden-draw}{RGB}{20,68,106} 
\definecolor{hidden-pink}{RGB}{255,245,247}

\usepackage{graphicx}%
\usepackage{multirow}%
\usepackage{amsmath,amssymb,amsfonts}%
\usepackage{amsthm}%
\usepackage{mathrsfs}%
\usepackage[title]{appendix}%
\usepackage{xcolor}%
\usepackage{textcomp}%
\usepackage{manyfoot}%
\usepackage{booktabs}%
\usepackage{algorithm}%
\usepackage{algorithmicx}%
\usepackage{algpseudocode}%
\usepackage{listings}%
\usepackage{natbib}

\begin{document}

\title[Article Title]{On Efficient Variants of Segment Anything Model: A Survey}


\author[1]{\fnm{Xiaorui} \sur{Sun}}
\author[2]{\fnm{Jun} \sur{Liu}}
\author[1,3]{\fnm{Hengtao} \sur{Shen}}
\author[1]{\fnm{Xiaofeng} \sur{Zhu}}
\author*[1]{\fnm{Ping} \sur{Hu}}\email{chinahuping@gmail.com}

\affil[1]{\orgdiv{School of Computer Science and Engineering}, \orgname{UESTC}}
\affil[2]{\orgdiv{ School of Computing and Communications}, \orgname{Lancaster University}}
\affil[3]{\orgdiv{School of Computer Science and Technology}, \orgname{Tongji University}}


\abstract{
The Segment Anything Model (SAM) is a foundational model for image segmentation tasks, known for its strong generalization across diverse applications. However, its impressive performance comes with significant computational and resource demands, making it challenging to deploy in resource-limited environments such as edge devices. To address this, a variety of SAM variants have been proposed to enhance efficiency while keeping accuracy. This survey provides the first comprehensive review of these efficient SAM variants. We begin by exploring the motivations driving this research. We then present core techniques used in SAM and model acceleration. This is followed by a detailed exploration of SAM acceleration strategies, categorized by approach, and a discussion of several future research directions. Finally, we offer a unified and extensive evaluation of these methods across various hardware, assessing their efficiency and accuracy on representative benchmarks, and providing a clear comparison of their overall performance. {To complement this survey, we summarize the papers and codes related to efficient SAM variants at \href{https://github.com/bhllx/On-Efficient-Variants-of-Segment-Anything-Model}{https://github.com/bhllx/On-Efficient-Variants-of-Segment-Anything-Model}}. 
}




\maketitle

\section{Introduction}
\label{sec:introduction}

The emergence of foundation models \cite{bommasani2021opportunities} has led to a thorough revolution in the field of artificial intelligence. Foundation models are large-scale neural networks pre-trained on massive data, which have powerful representing ability and strong generalization to perform various tasks. In the field of natural language processing, the recently popular research trend is about the large language models (LLMs) \cite{zhao2023survey, wan2023efficient}, which has witnessed great development with prominent works like OpenAI's GPT series \cite{NEURIPS2020_1457c0d6, achiam2023gpt}, Google's PaLM series \cite{chowdhery2023palm, anil2023palm} and Meta's LLaMA series \cite{touvron2023llama, touvron2023llama2}. Meanwhile, the success of Vision Transformer (ViT) \cite{dosovitskiy2020image} which firstly introduces the Transformer \cite{NIPS2017_3f5ee243} architecture into the field of Computer Vision, has brought up a new era for vision foundation models (VFMs). Vision-language foundation models like CLIP \cite{radford2021learning},  LLaVA~\cite{liu2024llava}, and Video-ChatGPT \cite{maaz2023video} etc., which aim at aligning the vision and language modalities, have demonstrated promising performance on plenty of downstream vision tasks\cite{li2024multimodal,gan2022vision,zou2024segment,tang2023video}. 

Recently, a novel foundation model for general image segmentation, the Segment Anything Model (SAM), was proposed by Meta \cite{kirillov2023segment}. Being fully trained on their proposed SA-1B dataset, which consists of more than one billion masks and eleven million images, with the task of achieving valid segmentation given any prompt (e.g. points, boxes, masks and text), SAM is able to well generalize to a wide range of downstream tasks (e.g. edge detection, object proposal and instance segmentation) with proper prompts. Shortly after its emergence, SAM has garnered significant attention from the research community and results in a surge of related works exploring SAM's generalization ability in various scenarios \cite{ma2024segment, chen2023sam, chen2023samfailssegmentanything}, including different image segmentation tasks \cite{roy2023sam, le2024medficientsam, ahmadi2023application, xie2023edit}, video analysis tasks \cite{zhang2023personalize, yang2023track, lu2023can, he2023scalable} and 3D vision tasks \cite{shen2023anything, dong2023leveraging, xu2024embodiedsam}. 
With the huge success of SAM, the upgraded Segment Anything Model 2 (SAM 2) \cite{ravi2024sam} is further proposed, aiming for efficient segmentation in both images and videos. SAM 2 introduces the streaming memory mechanism to extend SAM's ability to videos. It is trained on both the SA-1B dataset and SA-V dataset which is their newly collected video segmentation dataset. As a result, SAM 2 naturally holds the powerful generalization and surpasses SAM when handling segmentation tasks \cite{shen2024interactive, yamagishi2024zero, Wang_2025_CVPR, yu2024sam,tran20242nd, liu2024lsvos}. 
A further step in unified segmentation has been taken by the lately proposed Segment Anything Model 3 (SAM 3)~\cite{carion2025sam3segmentconcepts, sapkota2025sam2}, which is built upon a image-level detector and a video-level tracker, with the goal of detecting, segmenting, and tracking all instances corresponding to the provided concept prompts in both images and videos.

Despite its success in a wide range of applications, the original Segment Anything Model, particularly SAM-H, faces significant limitations due to its slow runtime and high computational cost. These challenges become even more pronounced when SAM is deployed in resource-constrained or real-time environments. As the demand for deploying machine learning models in practical, resource-constrained scenarios continues to grow, SAM's current design proves inefficient for widespread use. This has led to a pressing need for more lightweight and efficient variants that can maintain the model’s powerful segmentation capabilities while addressing these constraints. As the community strives to overcome these obstacles~\cite{zhao2023fast,varadarajan2023squeezesam,zhang2023faster,shu2023tinysam,wang2024repvit, zhou2023edgesam,lv2024ptq4sam,chen20230}, a comprehensive understanding of the latest advancements in making SAM more efficient is crucial. Consequently, it is both timely and necessary to conduct a detailed survey of the ongoing efforts aiming at improving SAM's efficiency and extending its applicability across diverse environments.

With the growing research related to SAM, several surveys have been presented to provide an overview from different perspectives~\cite{zhang2023survey, zhang2023comprehensive,zhang2023segment, zhang2024segment,zhang2024segment-videos,zhang2024unleashing, jiang2025prompt}. However, these existing surveys primarily focus on SAM's downstream applications and exhibit several limitations: 1) None of them address the emerging field of improving SAM's efficiency, which is gaining significant attention and is crucial for practical deployment in real-world applications. 2) With the exception of \cite{zhang2024segment-videos}, most surveys lack a structured taxonomy that would allow for clearer organization and reference. 3) Most prior surveys focus on collecting and describing SAM-based methods but lack a systematic evaluation or comparison of these approaches. To address these gaps, we conduct this survey, not only to comprehensively review the development of efficient Segment Anything Models, but also to fairly evaluate and compare them. The main contributions of this work can be summarized as follows:

\begin{itemize}
    \item{We provide a systematic review of the efficient SAM variants designed to accelerate segmentation tasks. We introduce a well-structured taxonomy for the methods, categorizing them based on the acceleration strategies they employ. To the best of our knowledge, this is the first survey focused specifically on this area.}
    
    \item{We offer comprehensive evaluations and comparisons of the efficiency and accuracy of these variants across different hardware, aiming at assisting researchers in selecting models that best meet their performance and application needs.}
    
    \item{We propose several potential directions for future research, offering insights that may inspire readers to contribute to the continued advancement of this field.}
\end{itemize}

The rest of this survey is organized as follows. In Section \ref{background}, we begin by introducing the background of the original SAM, followed by a review of model acceleration techniques that can be applied to enhance SAM's efficiency. In Section \ref{methodologies}, we categorize the existing methodologies based on their objectives and techniques, providing a detailed review of each category. We also discuss several potential research directions for further accelerating SAM. In Section \ref{evaluation}, we conduct unified evaluations of these models in terms of efficiency and accuracy. Finally, in Section \ref{conclusion}, we provide a summary of this survey. 

\begin{table*}[ht]
  \centering
  \small
  \caption{Efficient SAM Variants: Backbone and Key Features.}
  \label{tab:efficient_sam_variants}
  \begin{tabular}{l p{3.2cm} p{8.2cm}}
    \toprule
    \textbf{Variant} & \textbf{Backbone} & \textbf{Key Features} \\
    \midrule
    FastSAM           & YOLOv8-Seg (CNN) & All-instance segmentation then prompt-guided selection. \\
    SqueezeSAM        & CNN             & U-Net based encoder-decoder architecture. \\
    EfficientSAM      & ViT-Tiny/Small  & Retrain lightweight ViT encoder with SAM leveraged. \\
    RMP-SAM           & Light CNNs      & Unified dynamic decoder with multi-task adapters. \\
    MobileSAM         & TinyViT (ViT)   & Encoder-only distillation from SAM. \\
    Light HQ-SAM      & TinyViT (ViT)   & Merge MobileSAM’s efficiency with HQ-SAM’s quality. \\
    ESAM              & EffiFormerV2 (ViT) & Holistic distillation from an expert model. \\
    NanoSAM           & ResNet (CNN)    & Knowledge distillation and TensorRT acceleration. \\
    PicoSAM2/3        & CNN             & Distilling from SAM2/3 with U-Net based structure. \\
    RepViT-SAM        & RepViT (CNN)    & ViT-inspired CNN design and efficient distillation. \\
    EdgeSAM           & RepViT (CNN)    & Prompt-in-loop distillation and embedding of priors. \\
    EfficientViT-SAM  & EffiViT (ViT)   & Apply linear ReLU attention to cut computation. \\
    FastSAM3D         & ViT-Tiny        & Extends SAM to 3D with sparse flash attention. \\
    SAM-Lightening    & ViT-Tiny        & Dynamic layer-wise distillation for faster inference. \\
    RWKV-SAM          & MBConv + VRWKV  & Integrate linear RWKV operation for efficiency. \\
    TinySAM           & TinyViT (ViT)   & Hierarchical sampling with full-stage distillation. \\
    PTQ4SAM           & ViT             & Post-training quantization with adaptive granularity. \\
    AHCQ-SAM          & ViT             & Hardware-friendly PTQ-based method. \\
    CAR-SAM           & ViT             & PTQ framework with cross-attention reconstruction. \\
    PQ-SAM            & ViT             & Group activations for efficient quantization. \\
    SlimSAM           & ViT             & Structured pruning using disturbed Taylor importance. \\
    SuperSAM          & ViT             & One-shot NAS combined with pruning. \\
    SparseSAM         & ViT             & Structured sparsification for attention and MLPs. \\
    SAMfast           & ViT             & Native PyTorch optimizations yield an 8× speedup. \\
    Lite-SAM          & Lite-ViT (Hybrid) & Multi-scale pooling and AutoPPN for prompt generation. \\
    MobileSAMv2       & ViT              & Object-aware prompt sampling. \\
    AoP-SAM           & ViT              & Adaptive prompt prediction with coarse-to-fine sampling. \\
    \midrule
    SAM2              & Hiera (ViT)     & Streaming memory for video segmentation. \\
    Q-SAM2            & Hiera (ViT)     & Quantization aware training with novel calibration. \\
    SurgicalSAM2      & Hiera (ViT)     & Similarity-based memory pruning strategy. \\
    EfficientTAM      & ViT             & Memory pooling and attention simplification. \\
    EdgeTAM           & RepViT (CNN)    & Perceiver-based memory token compression. \\
    EfficientSAM2     & Hiera (ViT)     & Object-aware window routing and memory selection. \\
    SAM2fast          & ViT             & 13x latency improvement with pure PyTorch techniques. \\
    TinySAM2          & RepViT (CNN)    & Spatial-temporal memory compression strategy.  \\
    \midrule
    SAM3              & Perception Encoder  & DETR-like detector and SAM2-like tracker. \\ 
    EfficientSAM3     & PE + Efficient ViTs & Progressive hierarchical distillation. \\
    DART              & ViT                 & Shared backbone and batched decoding.\\
    Mix-QSAM3         & Perception Encoder  & Mixed-precision PTQ framework.\\
    \bottomrule
  \end{tabular}
\end{table*}

\section{Preliminaries} 
\label{background}
In this section, we will first introduce the details and applications of the Segment Anything Model (SAM) in Section \ref{sec: 2.1}. In Section \ref{sec: 2.2}, we briefly review efficient backbones that could potentially replace SAM's image encoder, and in Section \ref{sec: 2.3}, we discuss model compression techniques that could be applied to SAM. For a more comprehensive understanding of efficient backbones and model compression methods, we recommend referring to surveys \cite{papa2024survey, xu2023survey}.

\subsection{Segment Anything Model}
\label{sec: 2.1}
Segment Anything Model \cite{kirillov2023segment} is a powerful foundation model in the image segmentation field, which can unify most image segmentation tasks by a fundamental segmentation task, i.e. the promptable segmentation task, with prompt engineering \cite{liu2023pre}. Another significant contribution of this project is the SA-1B dataset, which has over 1B masks from 11M licensed and privacy-preserving images. Trained with such abundant and high-quality data, SAM is expected to have strong robustness and generalization. The huge potential of SAM has quickly sparked researcher's interest in both exploring its ability in a wide range of real-world applications and improving its architecture to segment more efficiently or accurately.

\begin{figure}[ht]
    \centering
    \begin{subfigure}{0.5\textwidth}
        \centering
        \includegraphics[width=1\linewidth]{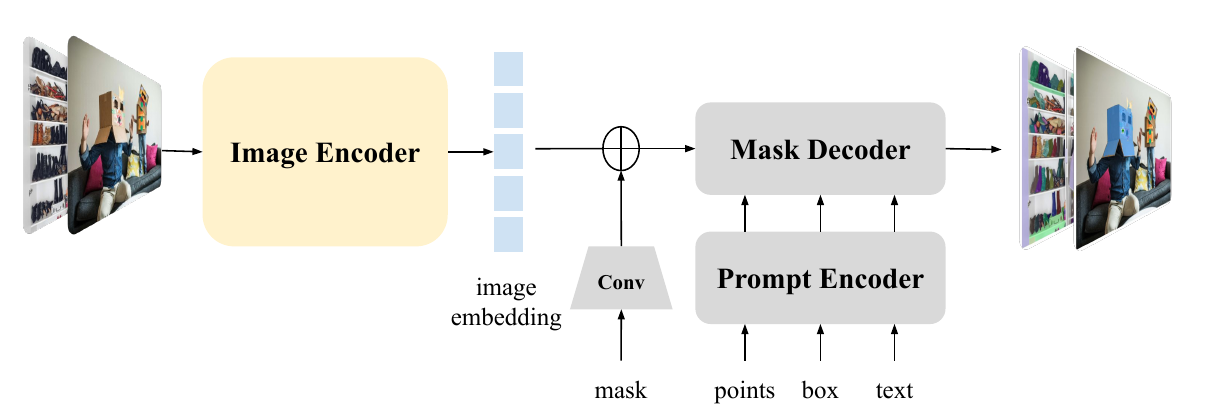}
        \caption{Segment Anything Model}
        \includegraphics[width=1\linewidth]{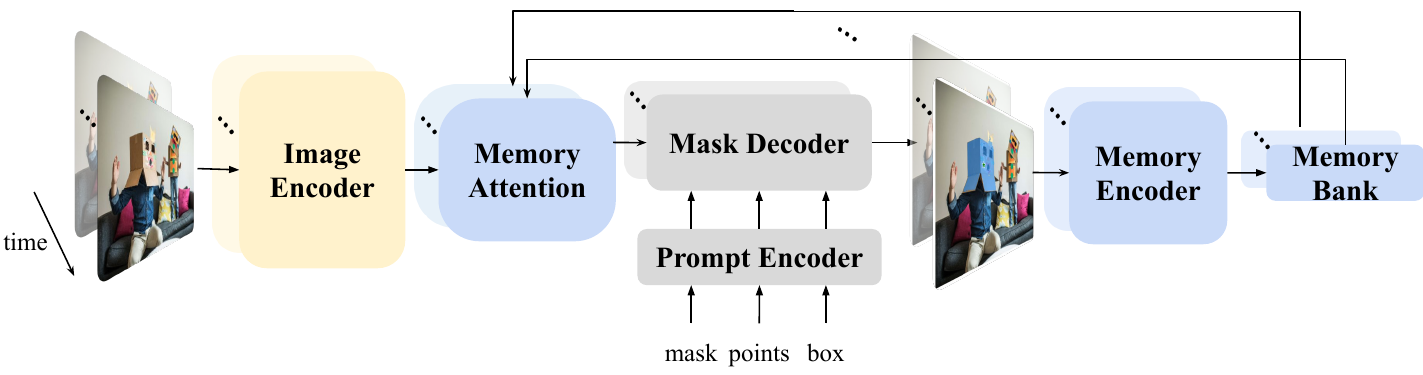}
        \caption{Segment Anything Model 2}
    \end{subfigure}%
    \caption{The architectures of (a) SAM  \cite{kirillov2023segment} and (b) the recent proposed SAM 2 \cite{ravi2024sam}. 
    }
    \label{fig:sam-arch}
    \vspace{-0.5cm}
\end{figure}

Recently, Segment Anything Model 2 (SAM 2) \cite{ravi2024sam} is proposed as a successor with a focus on efficient promptable visual segmentation (PVS) for both images and videos. To enable SAM 2 to segment anything in videos, researchers introduce the streaming memory mechanism into SAM's original architecture. SAM 2 is trained from scratch by two stages: 1) Pre-training on SA-1B dataset with the promptable segmentation task for images; 2) Training on mixed data with the promptable segmentation task for images and videos. Similar to SAM, researchers build up a data engine to generate a large-scale dataset for video segmentation, named the SA-V dataset. With both manually annotated and automatically generated masklets (object segmentation in videos), SA-V finally collects 642.6K masklets across 50.9K videos. In this survey, we consider SAM 2 as an efficient SAM variant and include it in the evaluation and comparison.

\subsubsection{Model} 
SAM consists of three components: an image encoder, a prompt decoder, and a mask decoder, as shown in Fig. \ref{fig:sam-arch} (a). The image encoder is an MAE \cite{he2022masked} pre-trained Vision Transformer(ViT) with minimal modification. It takes the preprocessed images as input and outputs an image embedding for each image. The prompt decoder is to embed prompts: points, boxes, masks, and text. The two embeddings are then fed into the lightweight mask decoder, which is built upon two modified Transformer decoder blocks \cite{NIPS2017_3f5ee243} and a few prediction heads, to generate valid masks. 
Based on SAM's architecture, SAM 2 additionally introduces a streaming memory mechanism. Specifically, a memory encoder, a memory bank, and a memory attention module. The structure of SAM 2 is illustrated in Fig. \ref{fig:sam-arch} (b). With the memory mechanism, SAM 2 can handle videos frame by frame. The memory encoder generates the memory of the current frame and sends it to the memory bank. The memory bank stores memories of recent predictions, feature maps of prompted frames, and high-level semantic information of the target objects (i.e. object pointers). The memory attention mechanism aims to make the image embedding from the image encoder fully interact with the information from the memory bank, resulting in refined embedding. Apart from the memory mechanism, SAM 2 also adopts the MAE pre-trained Hiera \cite{ryali2023hiera} as the image encoder which is more efficient than ViT-H, with expectation for faster speed.


\begin{figure}[t]
    \centering
    \includegraphics[width=1\linewidth]{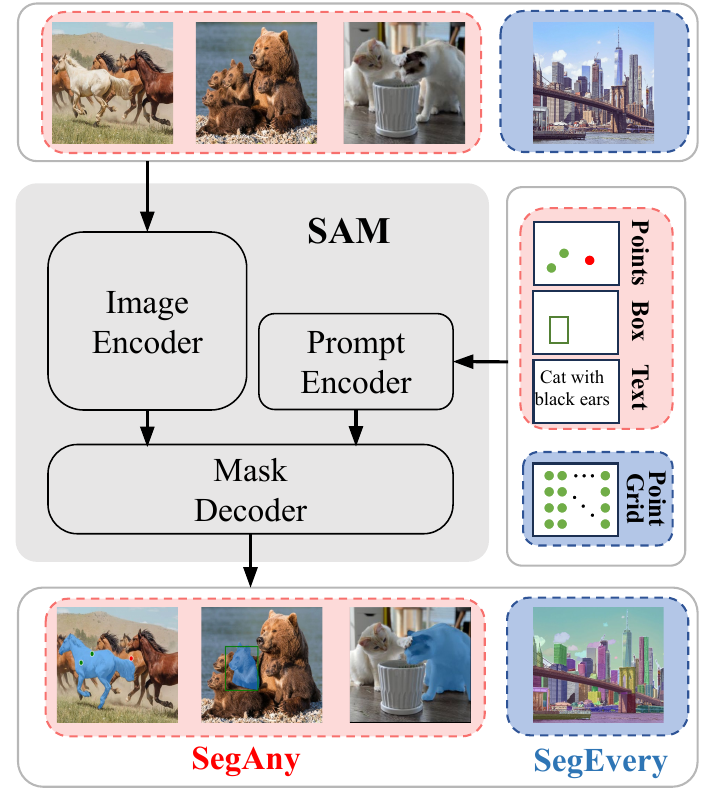}
    \caption{Illustration of the Segment Anything task (SegAny) and the Segment Everything task (SegEvery).}
    \label{fig:sam-app}
    \vspace{-0.5cm}
\end{figure}

\subsubsection{Task}
\label{sec: 2.1.1}
The promptable segmentation task is proposed as the fundamental task of SAM, whose goal is to return a valid mask with any given prompt (e.g. a point, a box, a mask, and text). This task is not only the objective during SAM's training, but also the base that enables SAM to solve various downstream tasks. Another important task is the all-mask generation which segments all objects in a picture. It is achieved by prompting SAM with a grid of points and then predicting masks based on these dense prompts. It is also a key procedure in the last stage of the data engine, which aims at enhancing the diversity of masks in SA-1B. As illustrated in Fig.~\ref{fig:sam-app}, the promptable segmentation task is called  \textbf{Segment Anything (SegAny)}, and the all-masks generation task is termed as \textbf{Segment Everything (SegEvery)}. These two tasks have summarized SAM's segmentation ability and have led to two research directions for enhancing SAM's efficiency. In this survey, We follow the two definitions, exploring SAM-based efficient variants' performance in both SegAny and SegEvery task.


\subsubsection{Application}
As SAM and its successor SAM 2 have demonstrated strong generalization in plenty of zero-shot downstream tasks \cite{kirillov2023segment, ravi2024sam}, the community is diving into exploring their application in more scenarios.  

One of the major applications of SAM is medical image segmentation. According to \cite{zhang2024segment}, works in this area can be classified into two types. Some aim at testing SAM's segmentation performance in CT images \cite{roy2023sam, hu2023sam, ayllon2025can}, MRI images \cite{mohapatra2023sam}, pathological images \cite{deng2023segment}, and so on \cite{sibley2026evaluating, li2025customized}. Others focus on improving the adaptation of SAM to these tasks by fine-tuning \cite{ma2024segment, li2024polyp}, auto-prompting \cite{hu2023skinsam, shaharabany2023autosam, Konwer_2025_CVPR} or framework modification \cite{zhang2023semisam}. 
{
Furthermore, as these SAM-based works also suffered from high latency and heavy computational cost, improving medical SAM methods' efficiency has increasingly gained attention in the community \cite{ma2024efficient, lv2026tipe}.
Several lightweight architectures \cite{le2024medficientsam, pfefferle2024daft, wei2024rep, gao2024swin, hu2025breastlightsam} have emerged, achieving both high-speed and high-quality segmentation across various medical imaging modalities, most of which distilled MedSAM's \cite{ma2024segment} image encoder into a lighter one to improve efficiency while maintaining the accuracy by subsequent fine-tuning. 
Optimization techniques during inference, for instance, exporting models to OpenVINO format and implementing an embedding caching mechanism \cite{le2024medficientsam, pfefferle2024daft}, could further speed up runtime efficiency, enabling variants to deploy on resource-constrained environments even without GPU. 
}

SAM is also applied to object detection across a diverse range of real-world scenarios \cite{zhang2023comprehensive, ji2024segment, guo2026learning, yang2026keeping, guo2026non, petti2025plantsam, camarena2025ad, ugwu2025promptable}, including crack detection \cite{ahmadi2023application} and crater detection \cite{giannakis2023deep} in civil infrastructure defect assessment, crop disease and pest detection \cite{li2023enhancing} in agriculture, anomaly detection \cite{cao2023segment} and remote sensing \cite{osco2023segment, wang2024samrs, Shan_2025_CVPR, wan2025systematic}. 
Moreover, Segment Anything has been adapted to Edit Everything \cite{xie2023edit}, Inpaint Anything \cite{yu2023inpaint}, and Caption Anything \cite{wang2023caption} to handle image editing tasks.
Additionally, SAM is applied to provide priors for high-quality multi-modality image fusion~\cite{Wu_2025_CVPR}.
Some researchers focus on exploring the uncertainty in SAM \cite{kaiser2025uncertainsam, Sheng_2025_CVPR}. For instance, UncertainSAM~\cite{kaiser2025uncertainsam} has proposed effective uncertainty quantization for SAM, and UNICL-SAM\cite{Sheng_2025_CVPR} further leverage the uncertainty to guide segmentation. 

Apart from image segmentation tasks, SAM has been widely applied to various video tasks \cite{zhang2024segment-videos}. A large portion of the work concentrates on two basic tasks: video object segmentation (VOS) \cite{zhang2023personalize, zhang2023uvosam, cheng2023tracking, Mei_2025_CVPR} and video object tracking (VOT) \cite{yang2023track, cheng2023segment, rajivc2023segment}. Researchers also explore SAM's application in tasks related to generation, for example, video super-resolution \cite{lu2023can} and video dataset annotation generation \cite{he2023scalable, delatolas2024learning}. Besides these, SAM is further utilized as an intermediate tool in video editing tasks \cite{zhao2023make, wu2023cvpr}.
Beyond 2D tasks, SAM is also extended to the 3D vision field. Shen et al. \cite{shen2023anything} applies SAM to 3D reconstruction and Dong et al. \cite{dong2023leveraging} applies it to 3D point cloud segmentation. The recent work \cite{xu2024embodiedsam} aims at achieving real-time segmentation for any 3D thing in an online setting.

For the subsequently proposed SAM 2, some studies have already explored its application in both image and video tasks \cite{xu2026segment, xu2026segment}. A popular trend is to apply SAM 2 to medical images and video tasks \cite{zhang2024unleashing}. Works like \cite{shen2024interactive, dong2024segment, yamagishi2024zero, xiong2026sam2} evaluate SAM 2's performance in medical images in both 2D and 3D modalities, while some others \cite{yu2024sam, lou2025zero} test it on surgical video segmentation and works like \cite{yin2025memory, xu2025tsms} attempt to improve SAM2's accuracy on this task. Researchers are also seeking strategies to adapt SAM 2 to medical tasks better \cite{yan2024biomedical, zhu2024medical}. Besides these, SAM 2 has been applied to some specific image segmentation tasks like digital pathology semantic segmentation \cite{zhang2024sam2}, mesh part segmentation \cite{tang2024segment}, solar panels segmentation \cite{rafaeli2024prompt}, and 3D instance segmentation~\cite{Zhao_2025_CVPR}. Moreover, several works \cite{adamyan2025samrl, ding2024sam2long, Videnovic_2025_CVPR, Cuttano_2025_CVPR, tran20242nd, liu2024lsvos} have improved SAM 2's performance on the VOT task, the Referring Video Object Segmentation (RVOS) task and the challenging Large-scale Video Object Segmentation (LSVOS) task. One popular trend is to optimize the memory mechanism, by capturing and integrating motion-aware~\cite{yang2024samurai, chen2025him2sam} or distractor-aware\cite{Videnovic_2025_CVPR} information.  


\subsubsection{Limitation}
Although SAM has demonstrated promising performance across various tasks, it still faces two key challenges in practical applications: 1) it often struggles to predict complete masks for fine structures, leading to imprecise boundaries; and 2) it is not real-time and remains resource-intensive, particularly when using heavy image encoders like ViT-H. To address these issues, some works aim to improve mask quality by utilizing high-resolution images~\cite{ke2024segment, huang2024hrsam}, integrating images and prompts~\cite{Yu_2025_CVPR}, or introducing optimized prompts~\cite{ Radman_2025_CVPR, Liu_2025_CVPR}, while others \cite{zhao2023fast, zhang2023faster, zhou2023edgesam, xiong2024efficientsam} focus on creating more efficient architectures to reduce SAM's time and resource consumption. 
Some recent works, such as ~\cite{wang2025tsda}, have also started to find a better balance between model's accuracy and efficiency. 
Previous surveys~\cite{ma2024segment, zhang2023uvosam, shen2023anything} have explored recent advances in enhancing SAM for higher-quality results. In this survey, we focus specifically on efforts to improve SAM's efficiency.

\begin{table}[!b]
    \centering
    \caption{Parameters of SAM's encoders. {Results in this table refer to~\cite{zhang2023faster}.}}
    \begin{tabular}{cccc}
    \toprule[1pt] 
        Parameters & SAM-H & SAM-L & SAM-B \\ \midrule
        Image Encoder & 632M & 307M & 86M \\ \midrule
        Prompt Encoder & \multicolumn{3}{c}{0.006M} \\ 
    \bottomrule
    \end{tabular}
    \label{encoder-size}
\end{table}

\subsection{Efficient Backbone}
\label{sec: 2.2}
SAM's inefficiency primarily stems from its heavy-weight image encoder. The sizes of SAM's image encoder are detailed in Tab. \ref{encoder-size}, with further estimates of SAM's total parameters provided in Section \ref{sec: 4.1}. As shown, in SAM-H, the ViT-H image encoder contains approximately 632M parameters, while the total model size is 641M, meaning the image encoder accounts for most of the model's parameters. Even in the smallest variant, SAM-B, the image encoder still has over 90$\%$ of the total parameters. Therefore, a straightforward yet effective method to accelerate SAM is to replace the large image encoder with more efficient backbones. These efficient backbones can include pure convolutional neural networks (CNNs) such as \cite{ronneberger2015u, he2016deep, wang2023repvit}, efficient vision Transformer architectures like \cite{wu2022tinyvit, li2022efficientformer, cai2022efficientvit}, and recent Transformer-alternative models like \cite{peng2023rwkv}. 


\subsubsection{Efficient Vision Transformer}
Efforts to improve the efficiency of Vision Transformers can generally be categorized into two approaches: 1) Designing more efficient architectures; 2) Refactoring the attention mechanism.

To reduce computational cost from a structural perspective, MobileViT \cite{mehta2021mobilevit}, a pioneering hybrid architecture, creatively integrates CNN blocks (MobileNetV2 blocks \cite{sandler2018mobilenetv2}) with Transformer blocks into a single model. Subsequent works such as \cite{wu2022tinyvit, li2023rethinking, cai2022efficientvit} basically follow this idea to build up efficient ViTs with hybrid structures, which have been widely used to substitute SAM's heavy image encoder. In \cite{zhao2023fast, shu2023tinysam}, TinyViT \cite{wu2022tinyvit} serves as the efficient backbone, while in \cite{Zhao2023ESAM} and \cite{zhang2024efficientvit}, EfficientFormerV2 \cite{li2023rethinking} and EfficientViT \cite{cai2022efficientvit} supplant SAM's original image encoder, respectively.
Another influential ViT design, MetaFormer \cite{yu2022metaformer}, which abstracts the attention mechanism into a broader concept called the token mixer, can deliver performance on par with Transformers using various token mixers. The simplest variant, PoolFormer, which uses pooling operations as the token mixer without introducing additional learnable parameters, has been leveraged as the base architecture to develop the light-weight image encoder for Lite-SAM \cite{fu2024lite}. 
The work \cite{liang2022expediting} expedites Vision Transformer by utilizing the creatively proposed token clustering and token reconstruction mechanism, which has already been applied to SAM's image encoder, resulting in the variant Expedit-SAM \cite{ExpeditSAM} with faster inference speed in image segmentation tasks.





Researchers have also made significant progress in optimizing the attention mechanism. It has been observed that the softmax operation within the attention mechanism contributes significantly to the overall computational cost. In EfficientViT \cite{cai2022efficientvit}, a novel ReLU linear attention is proposed to achieve a global receptive field with higher efficiency. This efficient backbone is further adopted in \cite{zhang2024efficientvit} to accelerate SAM. 
Improvements to the attention mechanism have also been made at the hardware level. FlashAttention \cite{dao2022flashattention} significantly reduces computation costs through techniques such as tiling, kernel fusion, and recomputation. And it is utilized in SAM acceleration works \cite{shen2024fastsam3d, songa2024sam} to reduce memory need and enhance computation efficiency.

\subsubsection{Transformer-alternative Models}    
\label{sec: 2.2.2}
While Transformers currently dominate both the language and vision fields, several newly proposed models have shown the potential to surpass them in terms of efficiency and performance.

State Space Models (SSMs), with near-linear computational complexity relative to sequence length, have emerged as promising alternatives to Transformers. Various methods \cite{gu2021efficiently, gupta2022diagonal, fu2022hungry} have been developed to enhance SSMs. One recent advancement, Mamba \cite{gu2023mamba}, has made significant strides in processing long sequences. Mamba's core innovation is the Selective State Spaces, which allow the model to dynamically decide whether to propagate or forget information, enabling more efficient sequence modeling. Additionally, the researchers incorporated a hard-aware algorithm to further accelerate computation on GPUs. Building on this, Vision Mamba \cite{zhu2024vision} applies Mamba to visual representation tasks. Vim introduces Bi-directional SSMs to better capture context within image patch sequences. With its strong performance and efficiency, Vision Mamba is emerging as a potential backbone for vision foundation models.

The Receptance Weighted Key Value (RWKV) model \cite{peng2023rwkv}, which combines the strengths of both recurrent neural networks (RNNs) and Transformers, achieves linear time complexity as sequence length increases. RWKV is well-suited for handling long sequence processing challenges. To facilitate global information interaction, RWKV replaces the traditional attention mechanism, which has quadratic complexity, with the more efficient WKV Operator and output gating mechanism. These are formulated as follows, 
\begin{align}
\label{eq:wkv}
wkv_t &= \frac{ \sum_{i=1}^{t-1} e^{-(t-1-i)w+k_i} \odot v_i + e^{u+k_t} \odot v_t }{\sum_{i=1}^{t-1} e^{-(t-1-i)w+k_i} + e^{u+k_t}}
\end{align}
\begin{align}
\label{eq:gate}
o_t &= W_o \cdot (\sigma(r_t) \odot wkv_t)
\end{align}
where \(r, k, v\) represents the shifted tokens of receptance, key, and value respectively, and \(W\) refers to weight.

RWKV has also been extended to the vision tasks. The Vision-RWKV (VRWKV) model \cite{duan2024vision} demonstrates performance comparable to Vision Transformers (ViT) but with greater efficiency. To adapt RWKV from 1D sequences to 2D images, Q-shift tokens is introduced to fuse neighborhood information in four directions. In \cite{yuan2024mamba}, a RWKV-based variant of SAM has achieved outstanding performance in efficiency, by adopting the efficient backbone mixed of MobileNetV2 blocks \cite{sandler2018mobilenetv2} and VRWKV blocks.


\subsection{Model Compression}
\label{sec: 2.3}

Model compression encompasses a range of techniques aimed at reducing the size and computational complexity of models, making it essential for deploying large models in real-world applications where computational resources are limited. The four primary methods for model compression and acceleration are knowledge distillation, quantization, pruning, and low-rank factorization.



\subsubsection{Knowledge Distillation}    
Knowledge distillation (KD) \cite{hinton2015distilling} was first introduced as a solution for deploying large, complex neural networks in resource-constrained environments. The core concept is to transfer the knowledge and representation capability from a larger, well-trained model (the teacher model) to a smaller, more efficient model (the student model).

When applying KD to accelerate SAM, the goal is to distill the knowledge from the original, larger SAM and impart it to more efficient SAM-like models. Given SAM's encoder-decoder architecture, KD can generally be categorized into two approaches: distilling the entire SAM model or distilling the image encoder alone. Most works, such as \cite{nanosam, zhang2023faster, wang2023repvit, shu2023tinysam, zhang2024efficientvit}, focus on distilling only the efficient backbone, while retaining the original SAM's prompt encoder and mask decoder. However, other approaches, like \cite{Zhao2023ESAM, zhou2023edgesam}, aim to distill the entire model by supervising the outputs of both the encoder and decoder.

{
\textbf{Discussion: }
Compared to training model fully from scratch, knowledge distillation usually requires fewer training resources, making it a relatively economical choice to build up lightweight and effective variants of large models, and thus being widely leveraged in accelerating SAM.
Moreover, with further fine-tuning on task-specific and smaller datasets, the distilled model can be effectively adapted to several downstream tasks, even in the resource-constrained environment which is not suited for deploying the heavy powerful  model.
However, distilling knowledge introduces additional supervision (i.e., the student model learns from both the soft labels of the teacher model and the ground truth labels of the task), making the training process more complex, which may lead to optimization difficulty and additional cost.
As a result, works accelerating SAM based on KD have all carefully designed the distilling pipeline, with expectation to improve the quality of distillation.
}



\subsubsection{Quantization}    
Quantization is the process to convert a model's high precision weights/activation values \(X\) (e.g. 32-bit floating point) to low precision formats \(X_q\) (e.g. 16-bit floating point, 8-bit integer). One of the widely used quantization functions, uniform symmetric quantization, is formulated as follows,
\begin{align}
\label{quant}
    X_q = Clamp(round(\frac{X}{s}), -2^b-1, 2^{b-1}-1) 
\end{align}
\begin{align}
\label{quant-reverse}
   X \approx \hat X =  X_q*s 
\end{align}
where \(b, s\) refers to the bit number in low precision and the scaling factor respectively, \(Clamp(x, a1, a2)\) clips integer \(x\) to \(a1/a2\) when \(x\le a1 /x \ge a2\). 
There are two primary types of quantization for neural networks: Post-Training Quantization (PTQ) and Quantization-Aware Training (QAT). PTQ applies quantization to fully trained models using a small calibration dataset to mitigate accuracy loss, while QAT adapts models to low precision during training, requiring the full training dataset. As a result, QAT is generally more costly and inefficient compared to PTQ.
Numerous PTQ methods have been developed for CNNs \cite{banner2019post, nahshan2021loss, wang2024aqa} and Transformer-based architectures \cite{frantar2022gptq, yao2022zeroquant, liu2021post, yuan2022ptq4vit}. In the context of accelerating SAM, works like \cite{shu2023tinysam, lv2024ptq4sam, zhang2025ahcqsam, wen2026car} utilize PTQ techniques to compress SAM with targeted strategies for improved efficiency.

{\textbf{Discussion: }
Post-Training Quantization is much more broadly investigated in lightening SAM, as it allows for directly applying quantization to the pre-trained foundation model with a few data for calibration, which is flexible and efficient, 
while on the contrary, Quantization-Aware Training requires re-training the model on entire training set, with heavier training cost.
However, PTQ-based light models usually suffer a bigger drop in accuracy compared to QAT-based ones, especially for those not explicitly trained to be quantization-aware.
In the practice of applying quantization on SAM while maintaining a reasonable level of performance, the extremely complex and asymmetric distribution rooted in SAM has made the process much more challenging.
}

\subsubsection{Pruning}    
\label{sec: 2.3.3}
Model pruning reduces a model's size and complexity by eliminating redundant weights or connections, while aiming to maintain accuracy as much as possible. Pruning methods are typically categorized into two types: structured pruning \cite{anwar2017structured, wang2019structured} and unstructured pruning \cite{laurent2020revisiting, chen2020tight}.

Structured pruning removes parameters in groups based on specific criteria, targeting substructures such as channels, layers, or blocks in a systematic manner. In contrast, unstructured pruning focuses on individual weights, often resulting in a sparse and fragmented network. However, unstructured pruning may not lead to effective acceleration on general hardware due to the irregularity of the remaining network structure. In \cite{chen20230}, structured pruning is applied to lighten SAM, significantly reducing the model's size while preserving most of SAM's capabilities by removing large amounts of redundant weights.

{\textbf{Discussion: }
Weights, channels and even layers in a large model are not equally important to contribute to the accuracy, making model pruning an intuitive yet effective method to reduce the inner redundancy. 
Nevertheless, the pruned model may not always achieve the expected speedup on certain hardware.
Another potential problem is that pruning aggressively can sometimes lead to the loss of generalization, which is unacceptable when lightening foundation models like SAM, leading to the demand for criteria that comprehensively evaluates weights' significance. 
}

\tikzstyle{my-box}=[
	rectangle,
	draw=hidden-draw,
	rounded corners,
	text opacity=1,
	minimum height=1.5em,
	minimum width=5em,
	inner sep=2pt,
	align=center,
	fill opacity=.5,
	line width=0.8pt,
]
\tikzstyle{leaf}=[my-box, minimum height=1.5em,
	fill=hidden-pink!80, text=black, align=left,font=\normalsize,
	inner xsep=2pt,
	inner ysep=4pt,
	line width=0.8pt,
]
\begin{figure*}[ht]
	\centering
	\resizebox{\textwidth}{!}{
		\begin{forest}
			forked edges,
			for tree={
                    grow=east,
                    reversed=true,
                    anchor=base west,
                    parent anchor=east,
                    child anchor=west,
                    base=center,
                    font=\large,
                    rectangle,
                    draw=hidden-draw,
                    rounded corners,
                    align=left,
                    text centered,
                    minimum width=4em,
                    edge+={darkgray, line width=1pt},
                    s sep=3pt,
                    inner xsep=2pt,
                    inner ysep=3pt,
                    line width=0.8pt,
				  ver/.style={rotate=90, child anchor=north, parent anchor=south, anchor=center},
			},
			where level=1{text width=16em,font=\normalsize,}{},
			where level=2{text width=18em,font=\normalsize,}{},
			where level=3{text width=18em,font=\normalsize,}{},
			where level=4{text width=18em,font=\normalsize,}{},
			where level=5{text width=12em,font=\normalsize,}{},
			[\textbf{Efficient SAM variants}, ver
                    [\textbf{Accelerating SegAny} (\S \ref{sec: 3.1}), fill=green!10
                        [\textbf{Training from Scratch}  (\S \ref{sec: 3.1.1}), fill=blue!10
                            [\textbf{SAM-different architectures}, fill=yellow!10
                                [\textbf{FastSAM \cite{zhao2023fast}}, fill=gray!10]
                                [\textbf{SqueezeSAM \cite{varadarajan2023squeezesam}}, fill=gray!10]
                            ]
                            [\textbf{SAM-like architectures}, fill=yellow!10
                                [\textbf{EfficientSAM \cite{xiong2024efficientsam}}, fill=gray!10]
                                [\textbf{RMP-SAM \cite{xu2024rap}}, fill=gray!10]
                                [\textbf{SAM 2 \cite{ravi2024sam}}, fill=gray!10]
                            ]
                        ]
                        [\textbf{Knowledge Distillation} (\S \ref{sec: 3.1.2}), fill=blue!10
                            [\textbf{With Light-weight ViT Encoders}, fill=yellow!10
                                [\textbf{MobileSAM \cite{zhang2023faster}}, fill=gray!10]
                                [\textbf{Light HQ-SAM \cite{ke2024segment}}, fill=gray!10]
                                [\textbf{ESAM \cite{Zhao2023ESAM}}, fill=gray!10]
                                [\textbf{TinySAM \cite{shu2023tinysam}}, fill=gray!10]
                            ]
                             [\textbf{With Pure CNN Encoders}, fill=yellow!10
                                [\textbf{NanoSAM \cite{nanosam}}, fill=gray!10]
                                [\textbf{PicoSAM2/3~\cite{bonazzi2025picosam2, bonazzi2026picosam3}}, fill=gray!10]
                                [\textbf{RepViT-SAM \cite{wang2024repvit}}, fill=gray!10]
                                [\textbf{EdgeSAM \cite{zhou2023edgesam}}, fill=gray!10]
                            ]
                            [\textbf{With Attention-modified Encoders}, fill=yellow!10
                                [\textbf{EfficientViT-SAM \cite{zhang2024efficientvit}}, fill=gray!10]
                                [\textbf{FastSAM3D \cite{shen2024fastsam3d}}, fill=gray!10]
                                [\textbf{SAM-Lightening \cite{songa2024sam}}, fill=gray!10]
                                [\textbf{RWKV-SAM \cite{peng2023rwkv}}, fill=gray!10]
                            ]
                        ]
                        [\textbf{Quantization} (\S \ref{sec: 3.1.3}), fill=blue!10
                            [\textbf{TinySAM \cite{shu2023tinysam}}, fill=gray!10]
                            [\textbf{PTQ4SAM \cite{lv2024ptq4sam}}, fill=gray!10]
                            [\textbf{AHCQ-SAM \cite{zhang2025ahcqsam}}, fill=gray!10] 
                            [\textbf{CAR-SAM \cite{wen2026car}}, fill=gray!10]  
                            [\textbf{PQ-SAM \cite{liu2024pq}}, fill=gray!10]      
                        ]
                        [\textbf{Pruning} (\S \ref{sec: 3.1.4}), fill=blue!10
                             [\textbf{SlimSAM \cite{chen20230}}, fill=gray!10]
                             [\textbf{SuperSAM \cite{abebe2025supersam}}, fill=gray!10]
                        ]
                        [\textbf{Code Refactorization} (\S \ref{sec: 3.1.5}), fill=blue!10
                             [\textbf{SAMfast \cite{SAMfast}}, fill=gray!10]
                        ]
                    ]
                    [\textbf{Accelerating SegEvery} (\S \ref{sec: 3.2}), fill=green!10
                        [\textbf{Panoptic Segmentation Strategy}, fill=blue!10
                            [\textbf{FastSAM \cite{zhao2023fast}}, fill=gray!10, tier=number]
                        ]
                        [\textbf{Efficient Sampling Strategy}, fill=blue!10
                            [\textbf{MobileSAMV2 \cite{zhang2023mobilesamv2}}, fill=gray!10, tier=number]
                            [\textbf{TinySAM \cite{shu2023tinysam}}, fill=gray!10]
                            [\textbf{LiteSAM \cite{fu2024lite}}, fill=gray!10]
                            [\textbf{AoP-SAM \cite{chenaop}}, fill=gray!10]
                        ] 
                    ]
                    [\textbf{Accelerating SAM2/3} (\S \ref{sec: 3.3}), fill=green!10
                        [\textbf{Backbone Optimization}, fill=blue!10
                            [\textbf{Q-SAM2 \cite{farronato2025q}}, fill=gray!10, tier=number]
                            [\textbf{DART \cite{turkcan2026detect}}, fill=gray!10]
                        ]
                        [\textbf{Memory Simplification}, fill=blue!10
                            [\textbf{SurgicalSAM2 \cite{liu2024surgical}}, fill=gray!10, tier=number]
                        ]
                        [\textbf{Two-pronged Approrach}, fill=blue!10
                            [\textbf{EfficientTAM \cite{xiong2024efficientsam2}}, fill=gray!10, tier=number]
                            [\textbf{EdgeTAM \cite{zhou2025edgetam}}, fill=gray!10]
                            [\textbf{EfficientSAM2 \cite{zhang2026efficient}}, fill=gray!10]
                            [\textbf{TinySAM2 \cite{ding2026tinysam}}, fill=gray!10]
                            [\textbf{EfficientSAM3 \cite{zeng2025efficientsam3}}, fill=gray!10]
                        ]
                        [\textbf{Quantization}, fill=blue!10
                             [\textbf{Mix-QSAM3 \cite{ranjan2026mix}}, fill=gray!10]
                        ]
                        [\textbf{Code Refactorization}, fill=blue!10
                             [\textbf{SAM2fast \cite{SAM2fast}}, fill=gray!10]
                        ]
                    ]
			]
		\end{forest}

 }
	\caption{Taxonomy of Efficient Variants of Segment Anything Model (SAM).}
	\label{fig: taxonomy}
    \vspace{-0.5cm}
\end{figure*}
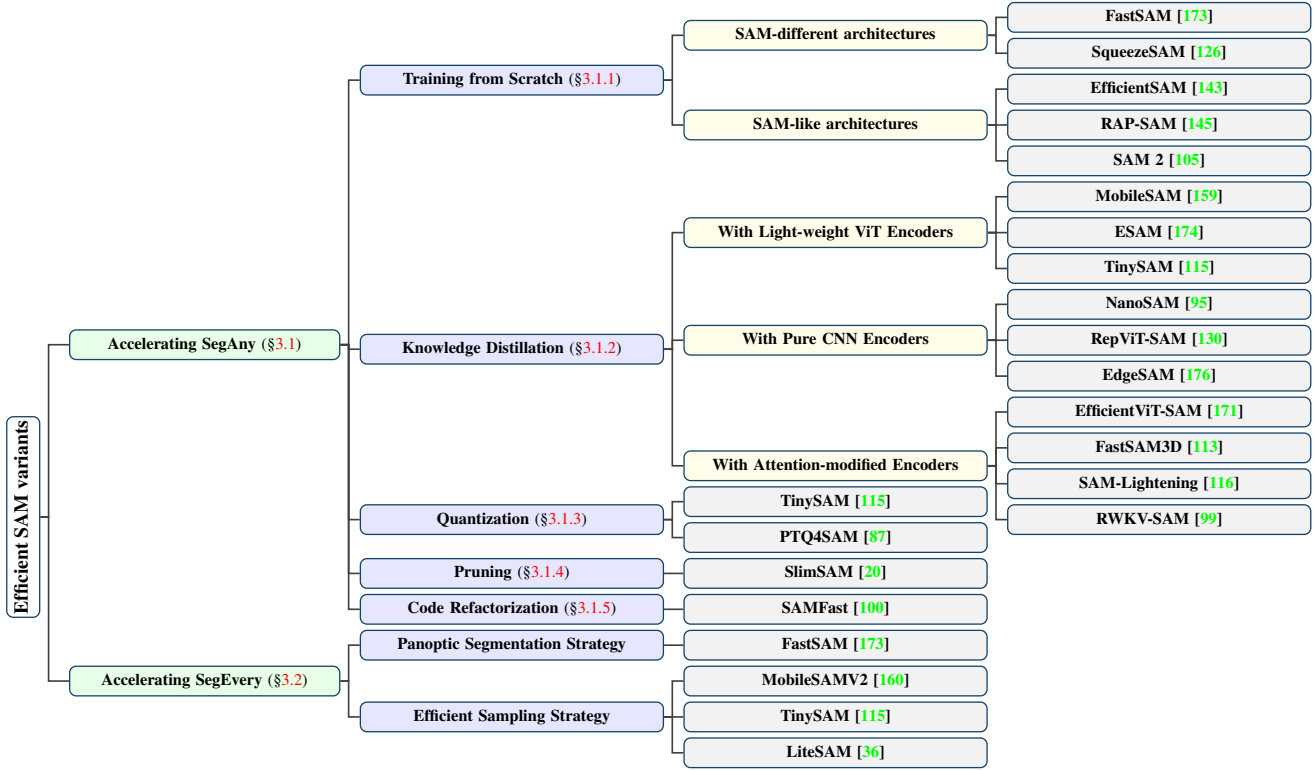
\subsubsection{Low-Rank Factorization}
Low-rank factorization is the technique that decomposes the high-dimension matrix into the product of low-dimension matrices, aiming at reducing the parameters of models for less space occupation and higher computational efficiency. A commonly used method of low-rank factorization is the singular value decomposition (SVD). It factorizes the complex matrix \(A \in \mathbb{R}^{m\times n}\) into three matrices and the calculation can be further rewritten as follows: 
\begin{align}
\label{low-rank-fac}
    A = U\Sigma V^T = \sum_{i=1}^r\sigma_i u_i v_i^T
\end{align}
where \(U\in \mathbb{R}^{m\times m},\ V\in \mathbb{R}^{n\times n}\) are orthogonal matrices and \(\Sigma\in \mathbb{R}^{m\times n}\) is a diagonal matrix with the none-zero singular values of \(A\), while \(r \leq \min\{m, n\}\) refers to the rank of \(A\). 
The low-rank factorization has been widely applied in deep neural networks, such as CNNs \cite{tai2015convolutional, yu2017compressing}, LSTMs \cite{grachev2017neural, winata2019effectiveness}, and Transformer-based models \cite{tahaei2021kroneckerbert, edalati2021kronecker, hajimolahoseini2022strategies}. However, to date, no studies have applied this technique to compress and accelerate SAM, presenting a potential future research direction.

{\textbf{Discussion: }
Sparse matrices derived from low-rank factorization, will take less memory usage and computational cost during forward pass, slimming and accelerating the vanilla model.
However, the implementation of some decomposition operations, for instance the SVD, yet being stable and effective in fully connected layers \cite{xue2013restructuring}, can be computationally expensive especially for large matrices \cite{swaminathan2020sparse}, which may hinders its application in SAM.
Moreover, determining the best matrix decomposition rank remains to be the crucial challenge for applying factorization approaches.
Furthermore, similar to pruning-based methods, decomposing matrices into low rank may result in loss of detailed information, undermining model's generalization performance.
Even though, some recent works applying low-rank factorization in LLMs \cite{yuan2023asvd, sharma2023truth, zhu2024survey}, exploiting the decomposition according to activation distribution or in specific Transformer layers, are proved to be effective, which may provide inspiration to researchers to further implement it on SAM.
}

\section{Efficient Variants of SAM}
\label{methodologies}
This section reviews the efforts to develop lightweight, efficient SAM-like models that have emerged since SAM gained prominence. These works aim to reduce the model's high computational cost and enable efficient performance, while preserving SAM's robust segmentation capabilities and generalization strength. As outlined in Section \ref{sec: 2.1.1}, SAM addresses two primary tasks including Segment Anything (SegAny) and Segment Everything (SegEvery). Accordingly, we discuss the research aimed at improving each task separately: Section \ref{sec: 3.1} focuses on accelerating the SegAny task, and Section \ref{sec: 3.2} covers efforts to accelerate the SegEvery task. Notably, some methods are applicable to both tasks, and we discuss these contributions individually.
{To help readers have a quick view of the efficient SAM variants, we provide Tab.~\ref{tab:efficient_sam_variants}, summarizing their backbone choices and key features.}
Additionally, we categorize all models into different classes based on the techniques they employ, and the taxonomy of methodologies is presented in Fig. \ref{fig: taxonomy}. Finally, in Section \ref{sec: 3.3}, we outline the potential directions for future research in this area.

\subsection{Accelerating SegAny Tasks}  
\label{sec: 3.1} 
 As analyzed in Section \ref{sec: 2.2}, the primary bottleneck of the SegAny task lies in SAM's heavy architecture. One straightforward solution is to replace the encoder with a more efficient backbone. Alternatively, adopting a different architecture that retains the same segmentation capabilities as SAM is another approach. Works following these strategies either involve training lightweight models entirely from scratch (discussed in Section \ref{sec: 3.1.1}) or training models using knowledge distillation with suitable supervision (covered in Section \ref{sec: 3.1.2}). Additionally, some research explores methods such as quantization, pruning, or localized optimizations to compress SAM directly, without replacing the encoder or constructing a new architecture. These efforts will be reviewed in Sections \ref{sec: 3.1.3}, \ref{sec: 3.1.4}, and \ref{sec: 3.1.5}, respectively.

\subsubsection{Training from Scratch} 
\label{sec: 3.1.1}
This subsection focuses on works that train SAM variants entirely from scratch. Based on their architecture, these models can be categorized into two types: SAM-different architectures and SAM-like architectures. We will explore each type in detail, following this classification.

\begin{figure}[t]
    \centering
    \includegraphics[width=1\linewidth]{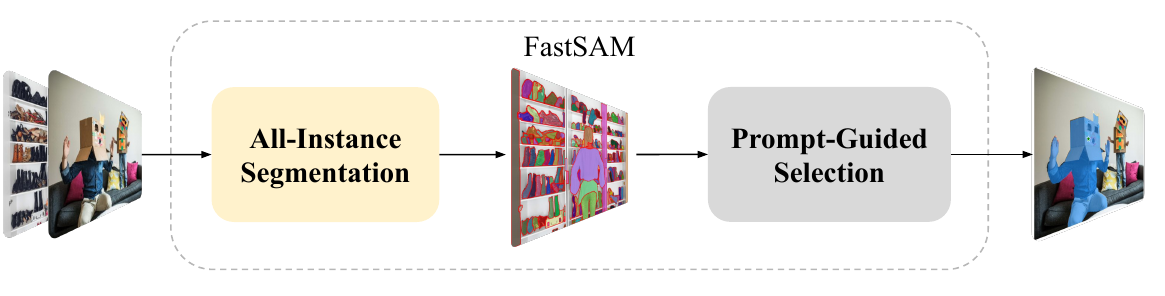}
    \caption{The architecture of FastSAM \cite{zhao2023fast}. It takes two stages to achieve segmenting anything: all-instance segmentation and prompt-guided selection.
    It is worth noting that the outputs of first stage are used directly as the SegEvery results.}
    \label{fig:fastsam}
\end{figure}

FastSAM~\cite{zhao2023fast} is one of the first SAM variants that does not rely on SAM's original Encoder-Decoder architecture. To achieve faster segmentation, it divides the SegAny task into two sub-tasks: all-instance segmentation and prompt-guided selection. Since instance segmentation has been effectively addressed by many CNN-based methods, FastSAM offers improved efficiency compared to the Transformer-based SAM. For instance segmentation, FastSAM employs the YOLOv8-Seg \cite{yolov8} model, using the YOLACT method \cite{bolya2019yolact} for enhanced performance. FastSAM can reliably predict objects of interest using points, boxes, or text as prompts.
In addition to accelerating the SegAny task, FastSAM also excels in the SegEvery task, as this can be efficiently achieved alongside all-instance segmentation. However, as an early efficient variant of SAM, FastSAM still has some limitations, such as producing low-quality masks for smaller objects and generating masks with less smooth boundaries. Despite these shortcomings, FastSAM marks significant progress by introducing a CNN-based architecture into this domain. The architecture of FastSAM is illustrated in Fig \ref{fig:fastsam}. 

\begin{figure}[t]
    \centering
    \includegraphics[width=1\linewidth]{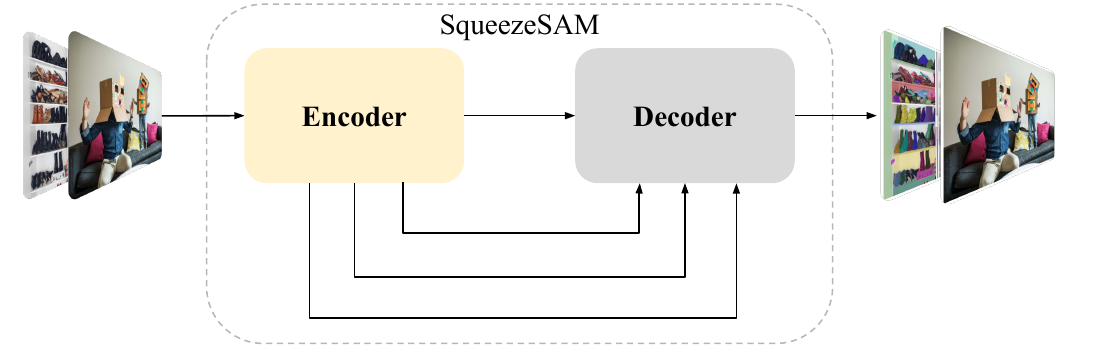}
    \caption{The architecture of SqueezeSAM \cite{varadarajan2023squeezesam}. It replaces the Transformer-based encoder-decoder structure with U-Net backbone\cite{ronneberger2015u}. Clicks from users and masks of salience objects are fed into SqueezeSAM with the input image to achieve interactive segmentation.}
    \label{fig:squeezesam}
\end{figure}

Building on the successful application of CNNs in SAM, Varadarajan et al. introduced SqueezeSAM~\cite{varadarajan2023squeezesam}, which further replaces SAM's Transformer-based architecture with a U-Net structure \cite{ronneberger2015u}. U-Net is composed of an encoder for feature extraction and a decoder for information recovery. SqueezeSAM retains the general U-Net architecture but incorporates two Transformer layers at the lowest scale of the U-Net to strike a balance between speed and accuracy.
Additionally, SqueezeSAM features several micro-level optimizations, such as capping the output channels at 256, using BatchNorm \cite{ioffe2015batch} instead of LayerNorm \cite{ba2016layer} for efficiency, and introducing skip connections \cite{he2016deep} between the encoder and decoder.
A unique challenge for SqueezeSAM lies in handling prompts. Unlike SAM, where prompt tokens are utilized at the decoding stage, SqueezeSAM adopts an early fusion strategy, adding encoded prompts as additional input channels before feeding them into the encoder. The model is trained from scratch using the SA-1B dataset, with data augmentation techniques addressing low-quality data issues.
SqueezeSAM is primarily designed for deployment in photography applications, where efficient interactive segmentation is needed. As depicted in Fig. \ref{fig:squeezesam}, its workflow involves generating an initial mask of the salient object, followed by fine-grained segmentation refined by user clicks.

\begin{figure}[t]
    \centering
    \includegraphics[width=1\linewidth]{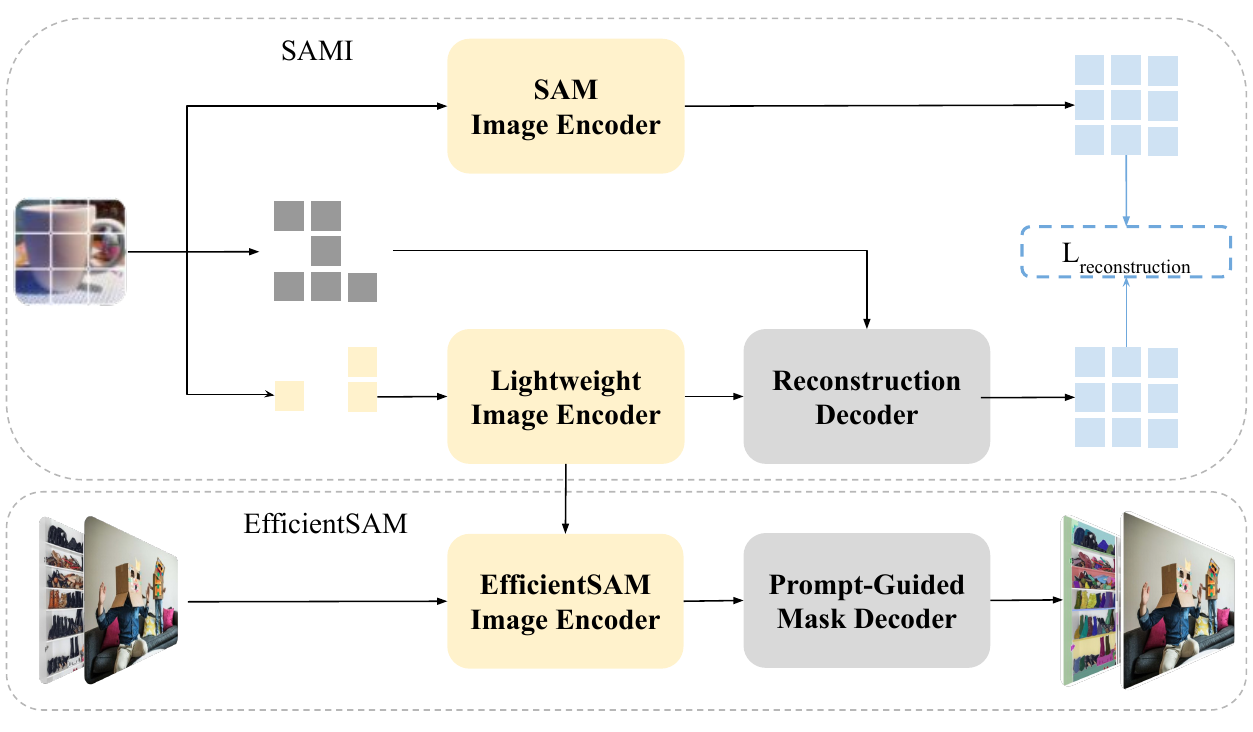}
    \caption{The framework of SAMI and efficientSAM \cite{xiong2024efficientsam}. The target of SAMI is to pretrain a light-weight encoder which holds representation ability close to SAM's image encoder. And the encoder is further extracted to build up the EfficientSAM.}
    \label{fig:efficientsam}
\end{figure}

{Different from FastSAM and SqueezeSAM, which introduce entirely new structures,} 
EfficientSAM~\cite{xiong2024efficientsam} retains SAM's original architecture but replaces the image encoder. They use ViT-tiny or ViT-small as a lightweight encoder and re-train them from scratch, leveraging the SAM-based Masked Image (SAMI) pretraining strategy. SAMI is adapted from the Masked AutoEncoder (MAE) framework \cite{he2022masked}, which was initially used to pretrain SAM's original image encoder.
SAMI follows an Encoder-Decoder pipeline: the encoder generates latent feature representations from unmasked patches, while the decoder reconstructs the missing embeddings of masked patches. This process is supervised by reconstruction loss, comparing the embeddings produced by SAM's ViT-H encoder with those generated by the SAMI pipeline. After pretraining, the lightweight encoder is extracted from the SAMI pipeline and integrated with the rest of SAM's components, forming EfficientSAM. The final step involves fine-tuning the entire model on the SA-1B dataset for further alignment and refinement.
SAMI is a general pretraining approach that can be applied to train any backbone for SAM variants. The overall structure of SAMI and EfficientSAM is illustrated in Fig. \ref{fig:efficientsam}.

\begin{figure}[t]
    \centering
    \includegraphics[width=1\linewidth]{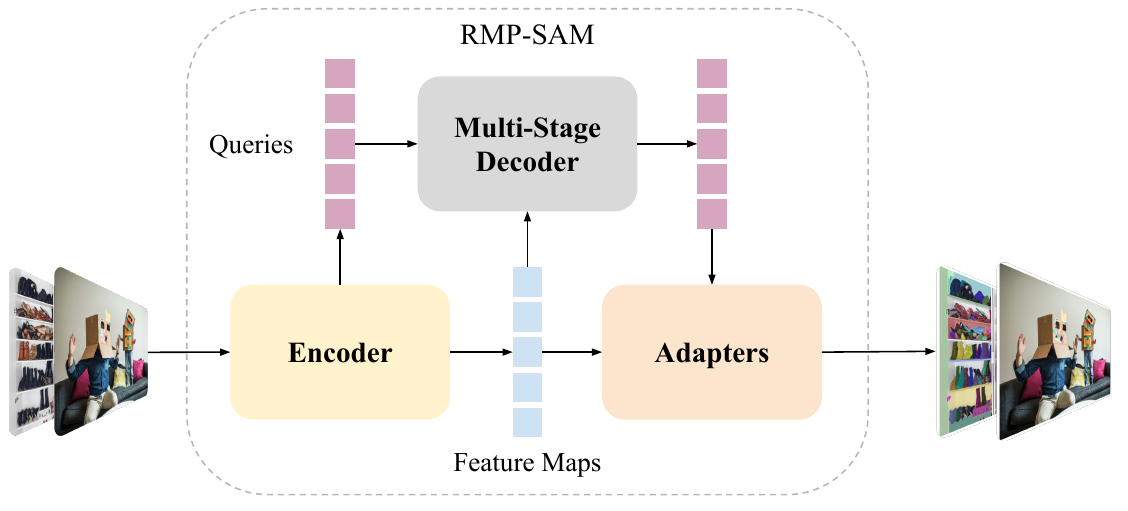}
    \caption{The architecture of RMP-SAM\cite {xu2024rap}. The encoder embeds images and videos to feature maps and encodes visual prompts as queries (tokens). The multi-stage decoder is to refine queries by interacting with feature maps. In the end, the feature maps and refined queries are fed into adapters to generate masks and classes.}
    \label{fig:rmpsam}
\end{figure}

{Concurrent with EfficientSAM, a new variant named RMP-SAM~\cite{xu2024rap}} 
is designed to achieve real-time, all-purpose segmentation, including panoptic segmentation (PS), video instance segmentation (VIS), and interactive segmentation (equivalent to the SegAny task). RMP-SAM retains SAM's basic Encoder-Decoder architecture but incorporates more efficient components to enhance performance.
For the encoder, RMP-SAM combines Feature Pyramid Networks (FPN) \cite{lin2017feature} with deformable convolutions \cite{dai2017deformable} to extract features from both images and videos, while using a prompt encoder to embed visual prompts. In the decoder, RMP-SAM employs a three-stage pipeline utilizing novel pooling-based dynamic convolutions to refine the mask tokens. The tokens generated in each stage, along with the feature maps from the encoder, serve as inputs. These inputs are first processed by dynamic convolutions \cite{chen2020dynamic}, followed by refinement using Multi-Head Self Attention (MHSA) and a Feed Forward Network (FFN).
After the decoder, two additional prompt adapters are introduced to enhance interaction between the visual prompts and segmentation tokens. The final masks are generated by computing the inner product between the updated tokens and updated prompts. The architecture of RMP-SAM is illustrated in Fig. \ref{fig:rmpsam}.

{Recently, Ravi et al. introduced the Segment Anything Model 2 (SAM 2) \cite{ravi2024sam}, an extension of the original SAM, with objective of delivering high-quality, real-time promptable segmentation across both images and videos.} In image segmentation tasks, SAM 2 is reported to achieve higher accuracy and a 6× improvement in efficiency compared to the original SAM. This significant advancement is largely attributed to its efficient image encoder, Hiera \cite{ryali2023hiera}, a hierarchical ViT that has been simplified from MViTv2 \cite{li2022mvitv2} by removing redundant components and utilizing the MAE framework for training. Hiera is a streamlined, purely transformer-based architecture that runs faster and delivers better accuracy than traditional ViTs in both image and video tasks.


\subsubsection{Knowledge Distillation based Methods}
\label{sec: 3.1.2}
From the taxonomy shown in Fig. \ref{fig: taxonomy}, we observe that many methods utilize knowledge distillation, as this approach typically requires less time and fewer resources compared to full model training. In this section, we review SAM variants that adopt efficient backbones for the image encoder while employing knowledge distillation for training. We categorize these models into three groups based on their encoder types: models with (i) lightweight ViT encoders,  (ii) pure CNN encoders, and  (iii) attention-modified encoders. We will introduce each category in turn.

\vspace{3pt}
\noindent\textbf{(i) Lightweight ViT Encoders} 
\vspace{3pt}

\begin{figure}[t]
    \centering
    \includegraphics[width=1\linewidth]{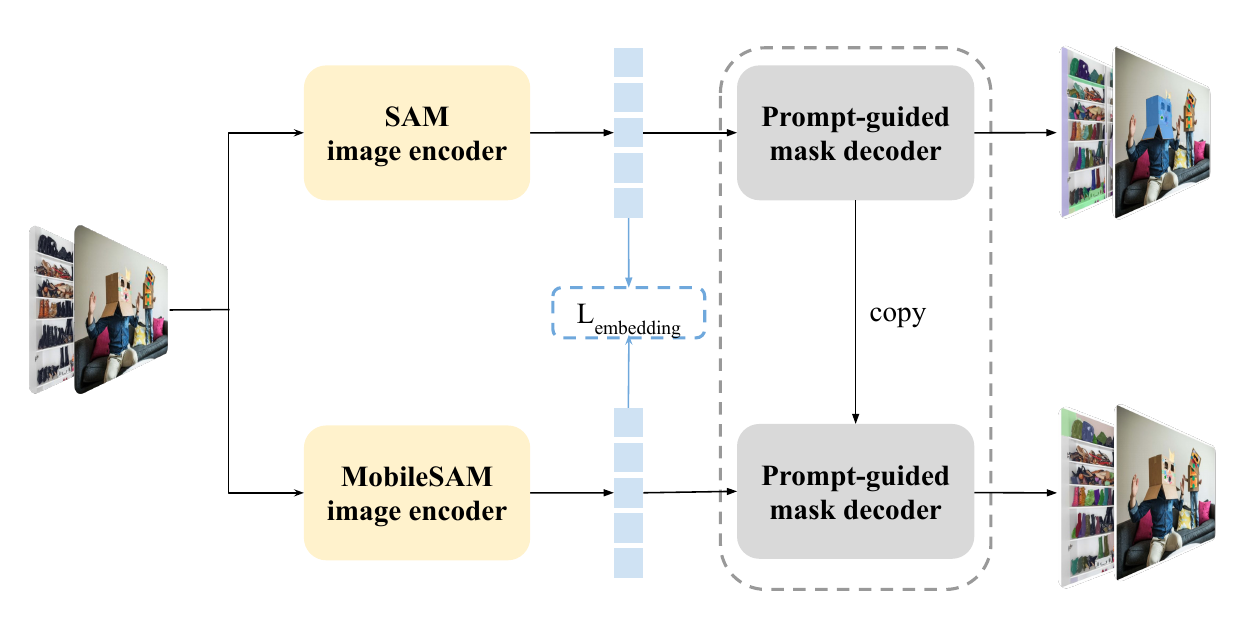}
    \caption{The framework of the encoder-only distillation for MobileSAM \cite{zhang2023faster}. The small ViT-based image encoder, TinyViT \cite{wu2022tinyvit}, is distilled from SAM's ViT-H and the prompt-guided mask decoder is inherited from SAM directly.}
    \label{fig:mobilesam}
\end{figure}

Zhang et al. \cite{zhang2023faster} made the first attempt to replace SAM's heavy ViT encoder with the more efficient TinyViT \cite{wu2022tinyvit}, resulting in the integrated model named MobileSAM . As highlighted in \cite{kirillov2023segment}, training SAM from scratch requires multiple days and 128 GPUs. MobileSAM attributes this complexity to the challenge of optimizing both the encoder and decoder simultaneously. To address this, they propose an encoder-only distillation strategy as illustrated in Fig. \ref{fig:mobilesam}, which aims to transfer the visual representation capabilities of ViT-H to TinyViT. The loss function used is a simple Mean Square Error (MSE) between the output image embeddings of the two encoders. Further fine-tuning of the prompt encoder or mask decoder is optional and can lead to improved accuracy.
The image encoder of MobileSAM has been further applied to HQ-SAM~\cite{ke2024segment}, a variant significantly boosting SAM’s mask quality, for mobile efficiency. The resulted model, namely light HQ-SAM, exceeds MobileSAM in performance with negligible gap in inference speed, proving it an excellent SAM variant trading off well between accuracy and efficiency.

Similar to MobileSAM, the later proposed ESAM \cite{Zhao2023ESAM} utilizes EfficientFormerV2 \cite{li2023rethinking} as its backbone, aiming for improved performance in CPU environments, particularly resource-constrained medical devices. Given that expert models \cite{wu2023medical, wang2022stepwise, chang2023esfpnet} often outperform SAM in medical applications, ESAM incorporates a novel Knowledge Distillation (KD) strategy called Holistic Knowledge Distillation (HKD) to transfer knowledge from an expert model to ESAM.
HKD involves two components: distillation on feature maps and distillation on output masks. For the feature map distillation, three different methods with varying focuses \cite{chen2021distilling, shu2021channel, yang2022focal} are combined to guide the learning process. For the output mask distillation, ESAM uses the Mean Square Error (MSE) loss between the teacher's and student's masks, supplemented by a Binary Cross-Entropy (BCE) loss between the teacher's masks and the ground truth masks. A Teacher Guided Module (TGM) is introduced, to further align the feature maps between the expert model and ESAM. 


\begin{figure}[t]
    \centering
    \includegraphics[width=1\linewidth]{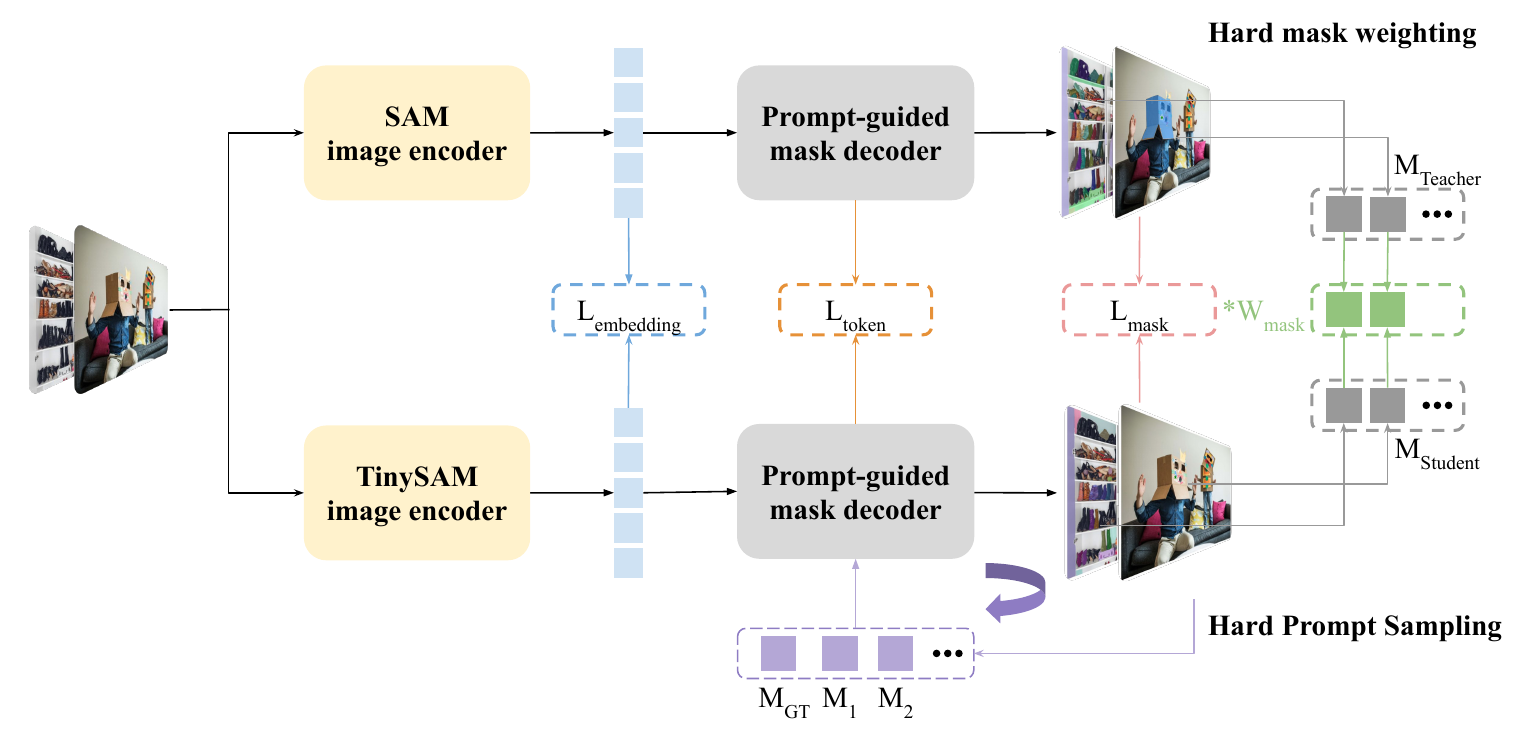}
    \caption{The framework of the hard mining full-stage knowledge distillation for TinySAM \cite{shu2023tinysam}. It contains two novel strategies to improve the quality of distillation: the Hard Mask Weighting and the Hard Prompt Sampling.}
    \label{fig:tinysam}
\end{figure}

{Not long after MobileSAM's publishing, Shu et al. \cite{shu2023tinysam} conducted an analysis on it,} 
identifying that encoder-only distillation can lead to significant performance degradation. To address this issue, they propose the Hard Mining Full-Stage Knowledge Distillation strategy for more effective distillation, as illustrated in Fig. \ref{fig:tinysam}. With a structure identical to MobileSAM, they developed a new SAM variant called TinySAM by training with this improved KD strategy.
Specifically, the strategy supervises not only the image embeddings but also the output tokens and output masks, all using L1 Loss. To further enhance the distillation process, they introduced the Hard Mask Weighting strategy, which assigns larger weights to masks that are harder to predict, thereby improving learning efficiency. The factor \(H\) is calculated as follows, 
\begin{align}
    H = sigmoid(\frac{IoU(M^T, M^{GT})}{IoU(M^S, M^{GT})+\epsilon}
\end{align}
where \(M^T, M^S, M^{GT}\) refers to the predicted masks of the teacher,
the student and the ground truth mask respectively. An extra strategy named Hard Prompt Sampling is adopted to sample prompts from areas with  prediction failure to make the model focus more on image's hard regions.

\vspace{3pt}
\noindent\textbf{(ii) CNN-based Encoders} 
\vspace{3pt}

Researchers from NVIDIA introduced a new SAM variant, NanoSAM \cite{nanosam}, based on MobileSAM. It is designed to achieve real-time performance on NVIDIA Jetson Orin platforms using NVIDIA TensorRT. NanoSAM replaces the ViT-based encoder with a pure convolutional network, specifically ResNet18 \cite{he2016deep}, while retaining the other components from MobileSAM. NanoSAM is distilled from MobileSAM, and both models are retrained with TensorRT for optimized performance.
The image encoder of MobileSAM is optimized with FP32 precision, whereas NanoSAM's image encoder is built using FP16 precision for faster execution. Inference latency results on Jetson Orin Nano and Jetson AGX Orin demonstrate that NanoSAM is 5x faster than MobileSAM, with minimal accuracy loss.
Very recently, a similar work, which builds up an ultra light-weight variant based on the U-Net architecture, named PicoSAM2~\cite{bonazzi2025picosam2}, has successfully deployed SAM on Sony IMX500 (a camera sensor with 8MB memory limit. 
Its successor PicoSAM3~\cite{bonazzi2026picosam3} further improves the performance by introducing architectural enhancements and employing distillation from SAM3.

{Due to MobileSAM with TinyViT as backbone still encounters challenges on resource-constrained devices,} Wang et al. \cite{wang2023repvit} developed an novel efficient SAM variant, RepViT-SAM, using their newly proposed CNN-based backbone, RepViT \cite{wang2024repvit}, as the image encoder. The core idea behind RepViT is to integrate the effective design principles of efficient Vision Transformers (ViTs) into lightweight CNNs. These design principles are applied at three levels: block-level, macro-level, and micro-level.
At the block level, RepViT separates the token mixer and channel mixer \cite{yu2023metaformer}, reduces the expansion ratio \cite{graham2021levit}, and increases the width of blocks. For the macro design, it incorporates early convolutions \cite{xiao2021early} as the input stem, deepens the down-sampling layers \cite{liu2022convnet}, employs a simpler classifier, and adjusts the block ratios across stages \cite{radosavovic2020designing}. At the micro level, only 3x3 convolutions are used, and Squeeze-and-Excitation layers \cite{hu2018squeeze} are applied only in odd-numbered blocks.
RepViT-SAM is trained by following the same distillation pipeline in \cite{zhang2023faster}, claiming to achieve a 10x increase in inference speed compared to MobileSAM.

 \begin{figure}[t]
     \centering
     \includegraphics[width=1\linewidth]{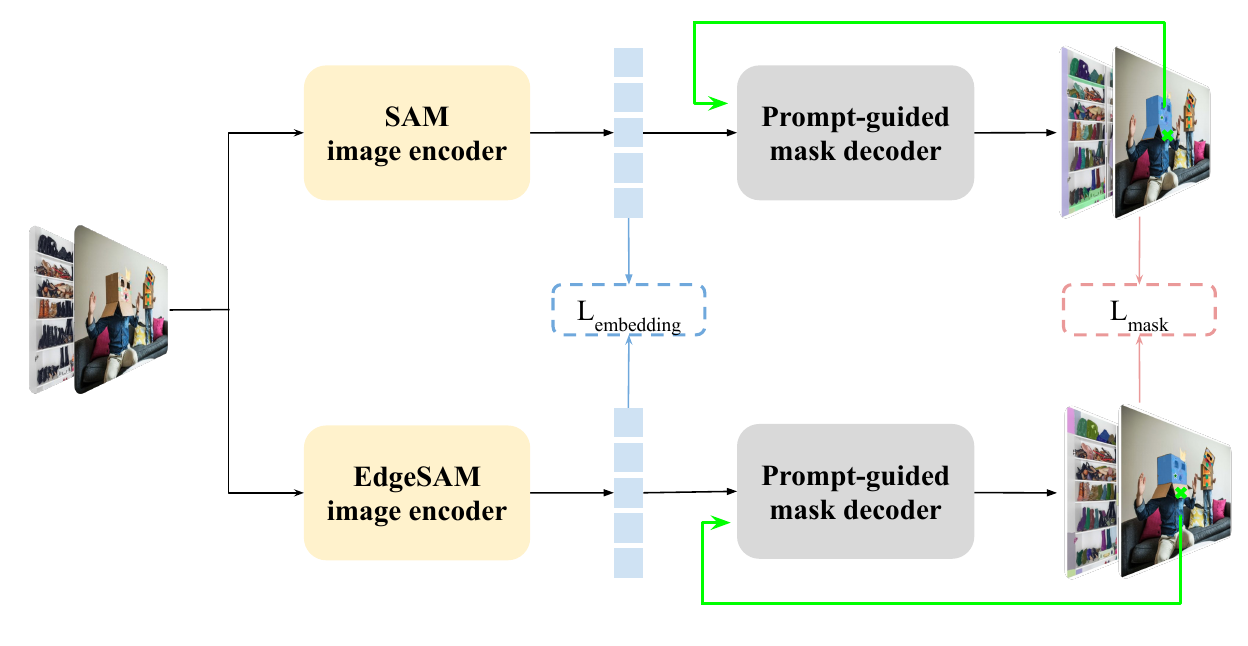}
     \caption{The framework of distilling EdgeSAM \cite{zhou2023edgesam}. It consists of the encoder-only distillation and the prompt-in-the-loop distillation.}
     \label{fig:edgesam}
 \end{figure}
Concurrent with the development of RepViT-SAM, Zhou et al. \cite{zhou2023edgesam} observe that MobileSAM \cite{zhang2023faster} still struggles to achieve real-time performance when deployed on edge devices, such as mobile phones. To address this, they introduced EdgeSAM, which replaces the Transformer-based encoder with the more lightweight and efficient, pure CNN-based RepViT, aiming for better performance on resource-constrained devices.
Similar to the approach in \cite{Zhao2023ESAM}, Zhou et al. argue that encoder-only distillation is inadequate, as it is task-agnostic and does not fully capture the model's task-specific needs. To overcome this, they propose the Prompt-in-the-Loop Distillation method, which adds additional supervision to the output masks. The "Prompt-in-the-Loop" refers to a dynamic sampling strategy that iteratively samples new prompts from the non-overlapping zones of the teacher's and student's predicted masks. After several iterations, the accumulated loss is backpropagated to update both the encoder and decoder.
To further enhance the output quality, EdgeSAM offers an optional module that embeds granularity priors from specific datasets. The overall framework for distilling EdgeSAM is illustrated in Fig.  \ref{fig:edgesam}.


\vspace{3pt}
\noindent\textbf{(iii) Attention-modified Encoders} 
\vspace{3pt}

{To further improve the efficiency of backbone while maintaining strong capability of feature extraction, Zhang et al.~\cite{zhang2024efficientvit} introduced EfficientViT~\cite{cai2022efficientvit} as image encoder, building up the powerful variant, EfficientViT-SAM.} The primary advantage of EfficientViT is its use of the ReLU linear attention mechanism, which facilitates global information interaction while enhancing hardware efficiency. By eliminating the hardware-unfriendly softmax operations and replacing them with ReLU, the attention calculation is reformulated as follows,
\begin{align}
    O_i = \frac{ReLU(Q_i)(\sum_{j=1}^NReLU(K_j)^TV_j)}{ReLU(Q_i)(\sum_{j=1}^NReLU(K_j)^T)}
\end{align}
once the sub-expressions \((\sum_{j=1}^N ReLU(K_j)^T V_j)\) and \((\sum_{j=1}^NReLU(K_j)^T)\) are computed, they can be reused for each query \(Q_i\). Therefore, the complexity is decreased to \(\mathcal{O}(N)\). The training of EfficientViT-SAM is divided into two steps. The first step is distilling from ViT-H to EfficientViT and the second step is train the entire model end-to-end on the SA-1B dataset. This approach achieves nearly no accuracy loss as well as huge improvement in efficiency. 

\begin{figure}[t]
    \centering
    \includegraphics[width=1\linewidth]{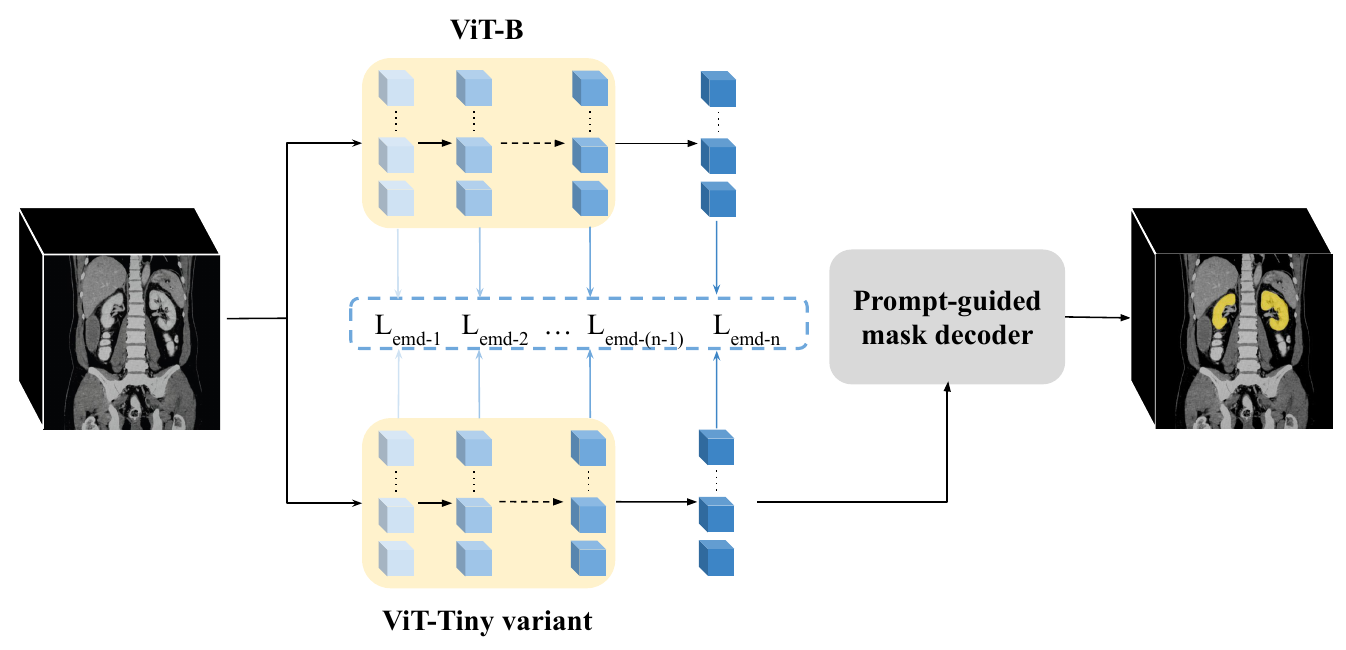}
    \caption{The framework of layer-wise progressive distillation of FastSAM3D \cite{shen2024fastsam3d}. It takes the ViT-B as teacher model to distill the efficient image encoder, a 6-layer ViT-Tiny variant. }
    \label{fig:fastsam3d}
\end{figure}

{Following the similar idea of applying high-efficiency attention mechanisms,} Shen et al. \cite{shen2024fastsam3d} proposed FastSAM3D, an efficient segment-anything model specifically designed for 3D volumetric medical images. The key contribution of this work is the development of the 3D Sparse Flash Attention mechanism. This novel attention approach combines the advantages of the 3D Dilated Attention \cite{ding2023longnet}, which extends the receptive field, with FlashAttention \cite{dao2022flashattention} to accelerate computations. FastSAM3D utilizes a modified ViT-Tiny as the image encoder, distilled from the ViT-Base encoder, ensuring efficiency without compromising performance. The authors implemented a layer-wise progressive distillation strategy to iteratively align feature maps between the two encoders, as shown in Fig. \ref{fig:fastsam3d}. 

Building on the approach of FastSAM3D, Song et al. \cite{songa2024sam} introduced a 2D variant called SAM-Lightening. Like its 3D counterpart, SAM-Lightening combines Sparse/Dilated Attention and FlashAttention as a replacement for the standard attention mechanism, while retaining the same image encoder. The key difference lies in the Knowledge Distillation (KD) strategy, referred to as Dynamic Layer-Wise Distillation (DLD). In DLD, a series of time-varying weights  \(\in [0, 1]\) are applied to determine which layers need to be updated and how much each layer contributes to the update process. As training progresses, the entire architecture is gradually optimized. SAM-Lightening reportedly achieves a 30× acceleration compared to SAM-H. The overall distillation framework is similar to FastSAM3D, with the main change being the substitution of the distillation strategy for dynamic layer-wise distillation.

A recent work by Yuan et al. \cite{yuan2024mamba}, RWKV-SAM, represents a significant advancement in accelerating SAM by introducing the popular linear attention model \cite{gu2023mamba, peng2023rwkv} as an efficient backbone. In their study, they compare RWKV-based and Mamba-based architectures and select the RWKV-based approach to construct a lightweight version of SAM. The backbone is a hybrid design, with the first two stages consisting of Mobile Convolution Blocks from \cite{sandler2018mobilenetv2}, and the final stage built using Vision RWKV blocks \cite{duan2024vision}. More details on RWKV can be found in Section \ref{sec: 2.2.2}.
Additionally, a refinement module is incorporated into the SAM-like architecture to enhance mask quality by fusing features from different levels generated at each stage. 
The model is trained using a "distillation-finetune" strategy, where knowledge from SAM-H is first distilled into the backbone, followed by fine-tuning of the entire model. RWKV-SAM demonstrates significant improvements in efficiency, while maintaining comparable segmentation performance to SAM.

\subsubsection{Quantization based Methods} 
\label{sec: 3.1.3}

As detailed in previous section, TinySAM \cite{shu2023tinysam} leverages the Hard Mining Full-Stage Distillation to improve the effectiveness of transferring knowledge. To further contract the scale of TinySAM, Shu et al. adopt a post-training quantization to the encoder of TinySAM and the quantified version is named as Q-TinySAM. They take the quantization basically following the instructions in \cite{yuan2022ptq4vit}. For the matrix multiplication in ViT, \( O=AB \) would be quantified to \(\hat O=\hat A\hat B\), with scaling factors \(s_A, s_B\). The researchers takes a alternative and iterative search strategy for the best \(s_A, s_B\) which can minimize the distance between \(O\) and \(\hat O\). The hessian guided metric is used to measure the distance which is approximately calculated as follows:
\begin{align} \label{eq: quant}
\min_{\Delta }\mathbb {E}[M^Tdiag\big ((\frac {\partial L}{\partial O^l_1})^2 ,\dots , (\frac {\partial L}{\partial O^l_{|O^l|}})^2\big )M]
\end{align}
where \(M=(\hat {O^l}-O^l)\) and the Kullback-Leible (KL) divergence of masks and IoUs are used for task loss.

Instead of taking quantization in an encoder-only manner used for Q-TinySAM,  Lv et al. \cite{lv2024ptq4sam} propose a novel framework that can directly take post-training quantization on SAM, namely PTQ4SAM. They start the work by first discovering the two challenges after traditional PTQ: 1) the existence of bimodal distribution which has negative effects on quantization quality and 2) the obvious discrepancy in the distribution of different attention mechanisms. Therefore, researchers bring up two strategies to solve them respectively: the Bimodal Integration (BIG) and the Adaptive Granularity Quantization (AGQ). For the Bimodal Integration, a sign factor \(\gamma\) is introduced to convert the bimodal distribution into a normal distribution. For the Adaptive Granularity Quantization, the key is to adopt an adaptive parameter \(\tau\) to adjust the base in the Log2 quantizer \cite{miyashita2016convolutional}. With the use of the AGQ strategy, the discrepancy between post-softmax distributions of different attention is expected to narrow down effectively. The PTQ4SAM is a plug-and-play SAM variant that can be easily deployed to downstream tasks.

{
Concurrent with PTQ4SAM, a counterpart named PQ-SAM~\cite{liu2024pq}, further exploits the post-training quantization method customized for SAM's activations, achieving the first usable 4-bit PTQ for SAM.
The core idea lies in transferring SAM's highly asymmetric activation distribution with extreme outliers into quantization-friendly distribution, for which the Grouped Activation Distribution Transformation (GAPT) is proposed.
GAPT firstly adopts tensor-wise outlier truncating to eliminate outliers in activation values, followed by the channel-wise outlier grouping to classify channels of truncated tensors into clusters based on their range.
Pairs of learnable shifting and scaling sizes are subsequently allocated to each channel group, reshaping the distribution as follows:
\begin{align} \label{eq: reshapeX}
\Tilde{X}(c) = (X(c) - {v_n}) \oslash {s_n}
\end{align}
where \(X(c), \Tilde{X}(c)\) represent the \(c^{th}\) channel of the original and reshaped activation tensor, \(v_n, s_n\) represent the \(n^{th}\) pair of shifting and scaling size and \(\oslash\) denotes for division operation.
With the transferred activations, a k-bit uniform quantization is further implemented, where the quantization step is jointly optimized with shifting and scaling size for optimal results.
}

\subsubsection{Pruning based Methods}

\label{sec: 3.1.4}
\begin{figure}[t]
    \centering
    \includegraphics[width=1\linewidth]{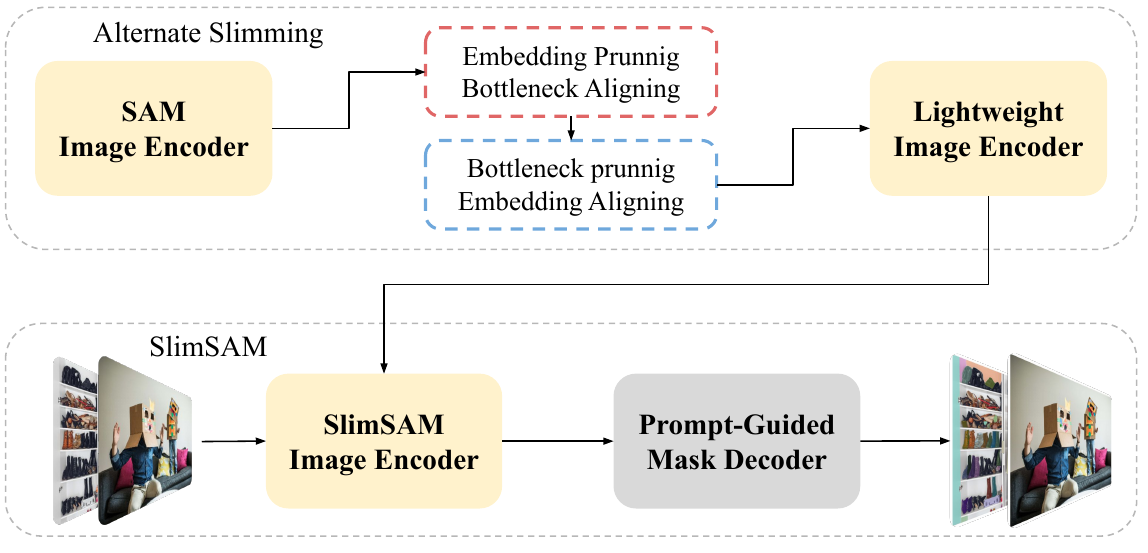}
    \caption{The overall pipeline of the alternate slimming strategy for SlimSAM \cite{chen20230}. It takes two steps to contract the heavy image encoder of SAM into a light-weight one: 1) Embedding pruning and bottleneck aligning; 2) Bottleneck pruning and embedding aligning.}
    \label{fig:slimsam}
\end{figure}
Chen et al. \cite{chen20230} were the first to develop an effective pruning strategy for reducing the size and complexity of SAM, resulting in the model known as SlimSAM. As discussed in Section \ref{sec: 2.3.3}, pruning algorithms aim to remove redundant parameters either structurally or individually. When applied to SAM's heavy encoder, the initial step involves estimating the importance of weights and activation values to determine which should be pruned. The core idea behind assessing importance is to evaluate the discrepancy in loss generated with and without the given parameter.
SlimSAM introduces the Disturbed Taylor Importance method, which uses the first-order Taylor expansion to approximate the importance of parameters and introduces Gaussian Noise \(\mathcal{N}\) to prevent gradients from becoming zero. This process is formulated as follows,
\begin {align} 
I_{w_i} & \approx \left | \mathcal L_{w_i}(x_i,t_i+\mathcal N)-\mathcal L_{w_i=0}(x_i,t_i+\mathcal N)\right | \nonumber \\
& \approx \left |\frac {\partial \mathcal L(x_i,t_i+\mathcal N)}{\partial w_i}w_i\right |
\end {align}
Once the importance of parameters is estimated, a strategy called Alternate Slimming is employed to perform structural pruning and post-alignment. The ViT-based encoder is first divided into two substructures: the embedding layer and the bottleneck layer. The strategy alternates between pruning the embedding/bottleneck layers to reduce model size and aligning the bottleneck/embedding layers to maintain model quality, ensuring both efficiency and performance. The workflow for this process is illustrated in Fig. \ref{fig:slimsam}.

\begin{figure}[t]
    \centering
    \includegraphics[width=1\linewidth]{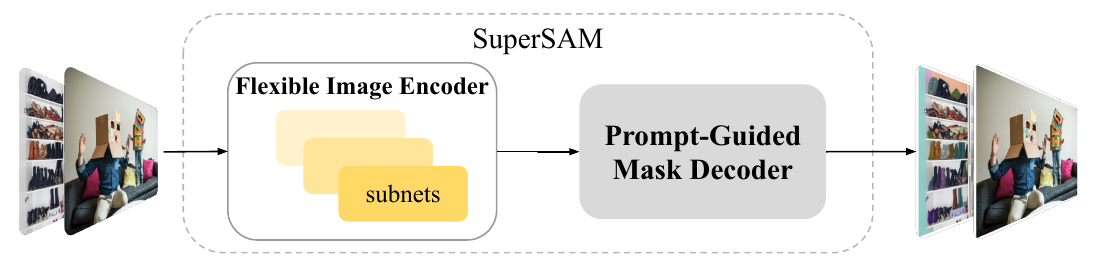}
    \caption{The architecture of SuperSAM~\cite{abebe2025supersam} consists of a flexible image encoder and SAM-like prompt-guided mask decoder. Lightweight SAM variants can be built up by sampling appropriate image encoder meeting predefined efficiency and accuracy requirements.}
    \label{fig:supersam}
\end{figure}

{
Unlike Chen et al.~\cite{chen20230} adopting the pruning and post-alignment strategy to lighten SAM in a straightforward manner, Abebe et al.~\cite{abebe2025supersam} incorporates pruning methods with the one-shot Neural Architecture Search~\cite{white2023neural} (NAS, an effective technique of automating the design of efficient neural networks), building up a weight-sharing supernetwork of SAM, named SuperSAM.
The overall architecture of SuperSAM is illustrated as Fig.~\ref{fig:supersam}.
SuperSAM works as a search space of SAM-like subnetworks with size 30-70\% smaller than original SAM-B, while some of them still maintaining considerable performance are considered as efficient variants of SAM.
The supernetwork is trained by sampling and training subnetworks with flexible image encoder, in which the structured pruning is applied to remove insignificant Transformer layers, with further weight reordering and slicing of MLP in remaining layers.
Once the SuperSAM is built up, techniques such as OpenTuner~\cite{ansel2014opentuner} can be deployed to search for the architecture meeting both efficiency and accuracy requirement.
The emerge of SuperSAM might further broaden the domain of investigating efficient SAM variants.
}

\subsubsection{Code Refactorization}
\label{sec: 3.1.5}
The Segment Anything Fast model (SAMfast), developed by the PyTorch team \cite{SAMfast}, is a rewritten version of SAM that leverages pure, native PyTorch optimizations. SAMfast is reported to be 8x faster than the original implementation while maintaining nearly the same accuracy. This improvement is the result of a systematic process of identifying bottlenecks and applying targeted optimizations.
Initially, the team identified long function calls that caused synchronous blocking, leading them to rewrite the corresponding code. Another significant bottleneck was the time-consuming matrix multiplication, which was mitigated by using bfloat16 precision. Following these adjustments, the team utilized torch.compile to fuse smaller operations and adopted PyTorch’s scaled\_dot\_product\_attention (SDPA) to accelerate attention computations on the GPU. Additionally, by integrating the new kernel built with Triton, memory usage on the GPU was further reduced.
When SAM uses the batch prediction method, input tensors of varying sizes are unified into NestedTensors, which significantly improve throughput. Despite these optimizations, matrix multiplications remained a key bottleneck. To address this, the team implemented int8 quantization and approximated matrix multiplication using semi-structured sparsity.
For more details on the step-by-step optimization process, it is recommended to go through the official blog for more details.




\subsubsection{Discussion}
\label{sec: 3.1.6}
In this part, we discuss the trade-offs among different architectures and methodologies, and suggest best practices for applying efficient SAM variants across various conditions. For SAM variants, the choice of backbone largely determines the model size, computational complexity, and segmentation performance. These backbones can be broadly classified into three groups: CNN‐based, ViT‐based, and attention‐modified architectures. CNN‐based backbones leverage small and parameter‐shared kernels for feature extraction, benefiting with fewer parameters and well-optimized hardware deployment, leading to efficient training and inference; however, their inherent locality limits the capture of global context and, consequently the segmentation accuracy. In contrast, ViT‐based structures overcome this limitation by employing attention mechanisms that enhance the context of feature representations and boost overall segmentation accuracy. This advantage comes at the cost of heavy training cost and quadratic computational complexity, rendering them computationally expensive and less efficient for reasoning. Attention‐modified backbones address the inefficiencies of traditional attention by either introducing more computationally efficient operations or by replacing the attention mechanism with Transformer‐alternatives, thereby reducing complexity while maintaining acceptable accuracy. However, efficient attention mechanisms may still suffer from challenges by hyperparameter sensitivity and implementation complexity.

{
Combining different architectural components within SAM variants might further overcome the above deficiencies by complimentarily taking advantages from each architecture, leading to better balance between efficiency and accuracy. 
Intuitively, hybrid designs that integrate CNN-based structures with Transformer-based modules or Transformer-alternative models are feasible solutions. There have been attempts in recent SAM variants, such as Lite-SAM~\cite{fu2024lite}, which employs a CNN-Transformer hybrid encoder  LiteViT, and RWKV-SAM~\cite{yuan2024mamba}, which integrates CNNs with the RWKV model. 
These variants with hybrid architectures have demonstrated significant improvements in efficiency while maintaining comparable segmentation performance to vanilla SAM.
Despite their promise, the mismatch between heterogeneous layers, which may lead to low quality in feature representations and high difficulty in model optimization, is a non-negligible challenge for building up effective hybrid models, and as a result, more advanced designs about module combination are needed in the future research.
}

In addition to backbone selection, additional compression techniques such as quantization and pruning can further reduce model size and memory usage during inference. However, these methods may not always deliver the expected runtime speedup and can sometimes lead to a significant degradation in accuracy and generalization. The training procedure itself also has a profound impact on performance. Training from scratch can yield comprehensive, high-quality models but requires substantial training resources. Knowledge distillation offers a more efficient alternative using less data. Yet, distilled models may fail to fully inherit the teacher’s capabilities, resulting in performance inferior to models trained from scratch. Moreover, combinations of these methods may yield models with varied accuracy–efficiency profiles. For example, pruning combined with post-quantization can compress a model to an extremely lightweight variant by reducing both parameter size and model complexity; subsequent fine-tuning (either through full training or distillation) can then help recover lost accuracy, thereby achieving a more balanced design. The recent work DC-SAM~\cite{zheng2025deep} achieves deep compression on SAM by first leveraging distillation and subsequently applying quantization.  

Based on the analysis above and the experimental results in Sections \ref{sec: 4.2} and \ref{sec: 4.3}, we suggest several best practices for deploying efficient SAM variants across diverse hardware platforms. In GPU environments, where computational resources are relatively abundant, thus throughput are the primary concerns. 
Models like EfficientViT-SAM offers an excellent balance between speed and segmentation accuracy. Its streamlined backbone featuring linear ReLU attention ensures the model capacity with reasonable memory cost, benefiting from the two-stage training procedure (initial distillation followed by full training) further enhances segmentation performance. Conversely, in CPU settings where parallel matrix computations are limited and models with high computational complexity can lead to prohibitive inference latencies, efficiency becomes critical. Architectures like NanoSAM with lightweight CNN-based backbones can theoretically minimize computational complexity. Our runtime evaluations on CPUs indicate that EfficientViT-SAM-L0 is the most effective variant, combining rapid inference with sufficient segmentation accuracy. For edge devices, which typically operate under stringent resource constraints and demand real-time performance, NanoSAM is the preferred option for minimizing memory usage and reducing latency. However, if the available hardware supports a slightly heavier model without compromising real-time requirements, EdgeSAM also emerges as viable alternatives that provide a more balanced combination of speed and accuracy.

\subsection{Accelerating SegEvery Tasks}  
\label{sec: 3.2}
As described in Section \ref{sec: 3.1}, the primary efficiency bottleneck for the SegAny task lies in the heavy image encoder. Any SAM variant with a more lightweight architecture will inherently be able to segment everything faster than the original SAM. However, as analyzed by Zhang et al. \cite{zhang2023mobilesamv2}, the main challenge of the SegEvery task stems from the dense grid sampling strategy. This strategy first predicts numerous masks based on a point grid and then selects valid masks, which is computationally expensive. Consequently, designing a more efficient sampling strategy to reduce the number of predicted masks has become the core approach to accelerating the SegEvery task.
Another potential solution is to convert the SegEvery task into another well-established task, such as all-instance segmentation, as done in FastSAM \cite{zhao2023fast}. In this part, we will review works that have specifically proposed optimized sampling strategies to accelerate the SegEvery task.

Building on the structure of SAM, Zhang et al. \cite{zhang2023mobilesamv2} introduced an object-aware prompt sampling strategy to enhance the efficiency of the SegEvery task. This project, named MobileSAMv2, operates independently of their previous work \cite{zhang2023faster} focused on accelerating the SegAny task. In MobileSAMv2, the researchers employ the YOLOv8 model, trained on a small subset of SA-1B, for object discovery. This model generates a large number of bounding boxes corresponding to latent objects. Highly overlapping boxes are filtered using Non-Maximum Suppression (NMS), and the remaining boxes are used as box prompts.
By using these filtered boxes as prompts, MobileSAMv2 eliminates the need to filter predicted masks—a much more time-consuming process. With the maximum number of prompts set to 320, the new strategy is reported to be 16 times faster than the traditional 32*32 grid sampling strategy. Additionally, MobileSAMv2 can be integrated with MobileSAM to create a unified model that achieves high efficiency in both SegAny and SegEvery tasks.

\begin{figure}[t]
    \centering
    \includegraphics[width=1\linewidth]{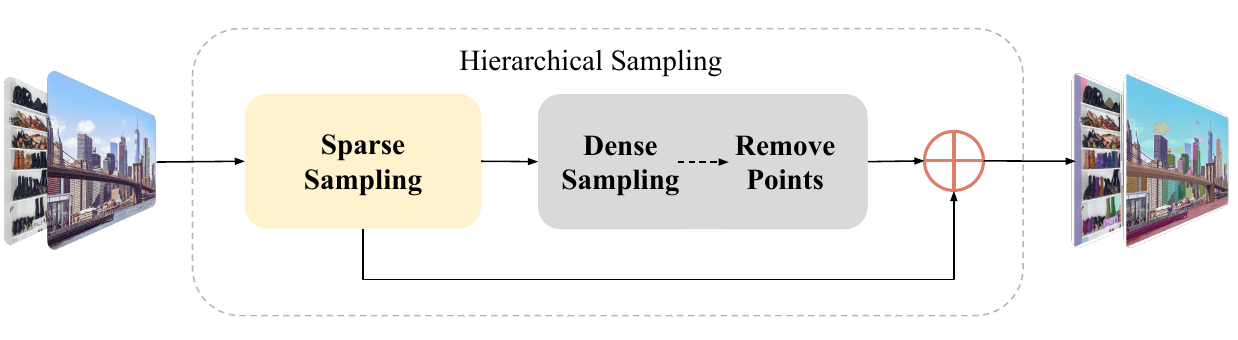}
    \caption{The pipeline of hierarchical sampling strategy for TinySAM \cite{shu2023tinysam}. The hierarchical strategy contains two rounds of sampling: one round for sparse sampling and the other round for dense sampling (including removing points from high confident regions). The final prediction comes from the fusion of results in two rounds.}
    \label{fig:tinysam-segevery}
\end{figure}

{Meanwhile, }Shu et al. \cite{shu2023tinysam} also observed that using a dense points grid (e.g., 32*32, 64*64) often generates a large number of redundant masks, which are later filtered out during post-processing, an operation that incurs significant time costs. In reality, only a few points within the grid are necessary to produce confident masks. To address this inefficiency, they proposed a hierarchical strategy for efficient sampling, which progressively selects optimal points for mask generation.
This strategy involves two rounds of prompt generation. In the first round, a sparse grid is used, incorporating only a fraction (approximately 1/4) of the default points along each side. Masks are generated based on these points, and after filtering, only high-confidence masks are retained as final predictions. In the second round, a denser grid is applied, following the default configuration. However, points located in regions already covered by confident masks are excluded, significantly reducing the number of points. The predicted results from both rounds are then fused to produce the final output.
The pipeline for this hierarchical strategy is illustrated in Fig. \ref{fig:tinysam-segevery}. By employing this two-round approach, the sampling process becomes both more time-efficient and fine-grained, leading to a notable acceleration in the SegEvery task with minimal performance degradation.

\begin{figure}[t]
    \centering
    \includegraphics[width=1\linewidth]{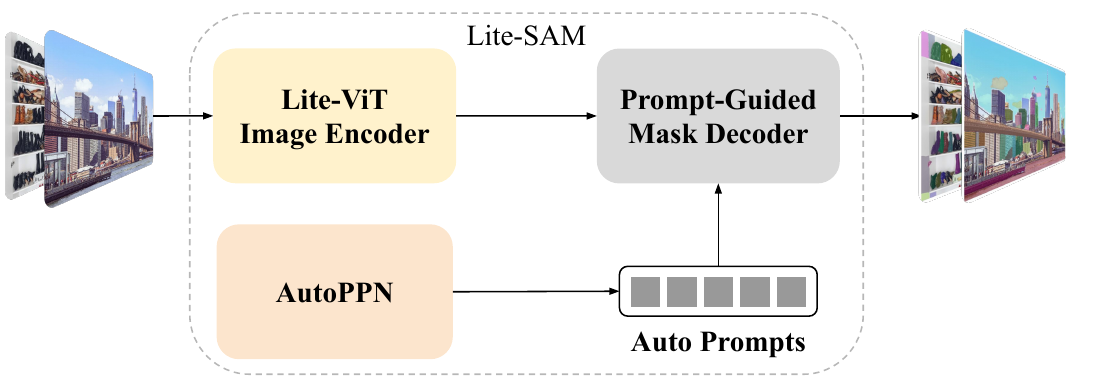}
    \caption{The architecture of Lite-SAM \cite{fu2024lite}. It takes the Lite-ViT as an image encoder and proposes an automated prompt proposal network (AutoPPN) for efficient sampling, while the prompt encoder and the mask decoder are copied from the original SAM.}
    \label{fig:litesam}
\end{figure}

Different from all aforementioned works, Fu et al. \cite{fu2024lite} propose an end-to-end training pipeline specifically designed for the SegEvery task, aiming to develop a SAM variant capable of segmenting everything more efficiently. Their model, named Lite-SAM, retains the overall architecture of the original SAM but replaces the heavy image encoder with a more lightweight solution. The architecture overview of Lite-SAM is shown in Fig. \ref{fig:litesam}.
Lite-SAM employs a CNN-Transformer hybrid structure called Lite-ViT, which consists of four stages with 2, 2, 6, and 2 Lite-ViT blocks, respectively. The key innovation in Lite-ViT is the Multi-Scale Pooling Module (MSPM), which serves as an alternative to the traditional attention mechanism. Adapted from the PoolFormer block \cite{yu2022metaformer}, MSPM utilizes channel-wise LayerNorm and extends the pooling operation to multiple scales.
As discussed earlier, another major bottleneck in SAM lies in the time-consuming grid sampling strategy. To address this, Lite-SAM introduces an Automated Prompt Proposal Network (AutoPPN) to improve sampling efficiency. AutoPPN takes the feature maps generated by the encoder as input and directly predicts point and box prompts. To ensure high-quality prompts, Lite-SAM uses a more powerful MSPM-based network instead of CNNs, and incorporates distance transform to estimate the confidence of point prompts.
While Lite-SAM is primarily designed to accelerate the SegEvery task, it also demonstrates improved efficiency in the SegAny task due to its lightweight image encoder.

\begin{figure}[t]
    \centering
    \includegraphics[width=1\linewidth]{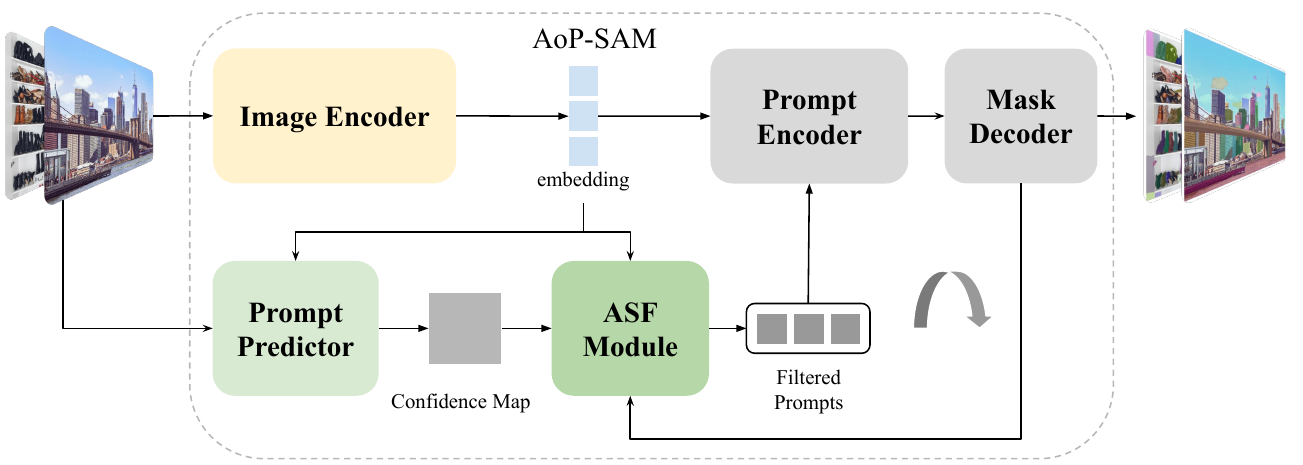}
    \caption{The architecture of AoP-SAM \cite{chenaop}. It adopts a Prompt Predictor to predict prompt confidence map and an Adaptive Sampling and Filtering (ASF) module to generate final prompts.}
    \label{fig:aopsam}
\end{figure}

{
Similar to Lite-SAM's idea of automatic prompt generation, the recent variant AoP-SAM proposed by Chen et al.~\cite{chenaop}, employs an efficient Prompt Predictor module to identify the optimal regions for prompt candidates, followed by an Adaptive Sampling and Filtering (ASF) module to generate prompts in a coarse-to-fine manner. 
The architecture of AoP-SAM is illustrated in Fig.~\ref{fig:aopsam}.
The Prompt Predictor consists of two encoder, processing original images and embedding from SAM's image encoder respectively, and one decoder, incorporating the spatial and context information from encoders to generate the prompt confidence map.
The function of ASF module is to extract essential prompts from prompt confidence map, which firstly samples the local maxima as coarse prompts, and secondly filters by the IoU scores of previously generated referenced masks and elimination scores of prompts, ensuring only crucial prompts will be retained for later mask generation.
The elimination score is calculated by referenced mask and image embedding, indicating the possibility of being redundant.
The experiments coudcted on three image segmentation datasets, have demonstrated AoP-SAM's highly efficiency for SegEvery task, surpassing both SAM's dense grid sampling strategy and MobileSAMv2's~\cite{zhang2023mobilesamv2} object-aware sampling strategy.
}

\subsection{Future Research Directions}
\label{sec: 3.3}
Through this comprehensive review of efficient SAM variants, we have outlined the current advancements in accelerating SAM. However, there remain opportunities for further exploration and innovation. Below, we present several potential future research directions, offering preliminary insights with the hope of inspiring readers to contribute to the ongoing development in this field.

\noindent\textbf{Exploring Advanced Architectures.}
While current SAM variants have demonstrated efficiency gains through the adoption of efficient architectures and model compression techniques, there is substantial potential for further improvement. The exploration of Transformer-alternative models, such as Mamba~\cite{gu2023mamba}, RetNet~\cite{sun2023retentive}, KAN~\cite{liu2024kan}, and TTT~\cite{sun2024learning}, presents an exciting opportunity to design more lightweight and efficient structures. 
{With the success of incorporating RWKV into SAM~\cite{yuan2024mamba}, leveraging Transformer-alternative models as backbone has shown great potential to build up efficient SAM variants, which may offer advantages in computational efficiency without sacrificing segmentation accuracy.}
In addition to alternative models, {refining attention modules' efficiency remains critical for further development of accelerating SAM, inspired by the preliminary exploration of FastSAM3D~\cite{shen2024fastsam3d}.}
Methods such as linear attention, low-rank factorization, or hybrid architectures combining convolutional and attention-based designs, which have been widely applied to Transformer-based models~\cite{Han_2023_ICCV, hajimolahoseini2022strategies, fang2022hybrid, schmidt2025segment, yang2025towards}, should be further investigated for efficient SAM.
Addressing bottlenecks in both computation and memory usage will enhance SAM's deployment across diverse hardware environments.

\noindent\textbf{Exploiting Sparsity and Acceleration Techniques.}
The inherent sparsity observed in deep neural networks, where only a subset of parameters significantly contributes to model output, offers a promising avenue for improving SAM's efficiency. 
The discovery in very recent work, SuperSAM~\cite{abebe2025supersam}, showing that the subnetwork with 7 layers in image encoder pruned (12 layers in total) can still maintain over 60\% performance compared to vanilla SAM, further provides proofs to SAM's sparsity, {which poses a promising research direction for further uncovering the fine-grained sparsity in SAM's architecture.}
Techniques like pruning, quantization, and structured sparsity {\cite{SAMfast}} could further reduce SAM's computational demands. {While several initial efforts in sparsification, such as those leveraged in SlimSAM~\cite{chen20230} and PQ-SAM\cite{liu2024pq}, have shown success, future research could dive deeper into understanding the distribution and dynamics of sparsity in SAM.} This includes investigating the optimal layers or components of SAM that can be pruned or sparsified without impacting performance {\cite{abebe2025supersam}}. 
Additionally, techniques such as sparse attention mechanisms, dynamic pruning during inference, and low-precision training, which have been deployed on Transformer-based architectures~\cite{lou2024sparser, rao2021dynamicvit, NEURIPS2023_20bd42d8}, can be further explored to balance accuracy and efficiency, especially for large-scale deployments.
The recent SparseSAM~\cite{tran2026sparsesam} applies a structured sparsification framework for attention and MLP layers, achieving 2x acceleration with negligible performance degradation. 
Moreover, Fan et al.~\cite{fan2025compress} proposes a novel compression technique, based on the Hyper-Compression principle, which projects high-dimensional parameters into scalars, providing new insights for lightening SAM.
By combining these with advanced knowledge distillation techniques, more compact, efficient variants of SAM could be realized.

\noindent\textbf{Hardware-Specific Optimizations.} Optimizing SAM for specific hardware platforms, including GPUs, TPUs, specialized AI accelerators (e.g., NVIDIA's TensorRT or Google's Edge TPU), and edge devices, can unlock significant gains in performance and efficiency. 
{Equipping SAM with hardware-aware model optimization techniques, for instance, operator fusion, quantization-aware training, and custom CUDA kernels, some of which have been applied in SAMfast~\cite{SAMfast} by PyTorch extensions, requires further exploration in the future to maximize throughput and reduce latency when deploying SAM on modern hardware platforms. }
In the context of edge devices, which face extreme constraints in storage, computational power, and energy supply, these optimizations are crucial for real-time applications, such as segmentation on UAVs or IoT devices. Future research could explore hierarchical cloud-edge architectures to offload computationally expensive tasks to the cloud, while running lightweight models locally on edge devices.
{This idea has already been investigated in LLMs for robot control~\cite{luan2023hierarchical}}.
Additionally, leveraging specialized AI hardware like Field-Programmable Gate Arrays (FPGAs) {\cite{zeng2024flightllm}} or using techniques such as hardware-aware neural architecture search (NAS) and mixed-precision quantization {\cite{xiao2023patch}} might further optimize SAM for low-latency, resource-constrained environments, ensuring that the model operates effectively across diverse hardware platforms. 
{Among them, NAS is now successfully applied to SAM, building up the first supernetwork of SAM~\cite{abebe2025supersam}.}


\noindent\textbf{Computation Cost Attacks and Defenses.} Substantial progress has been made by the research community to improve the efficiency of deep learning models, particularly for deployment on resource-constrained edge devices. However, recent studies have exposed vulnerabilities in some models when subjected to computation cost attacks, which deliberately increase resource consumption and processing time far beyond normal operational levels \cite{hong2020panda, chen2022nicgslowdown, chen2023dark, gao2024inducing, pan2022gradauto}. Despite these findings, the efficiency and robustness of Segment Anything Model (SAM) variants under computation cost attacks remain largely unexplored, especially when techniques like dynamic models are applied. This presents a valuable opportunity for future research. One direction could involve identifying specific vulnerabilities in SAM variants when exposed to such attacks. Additionally, the insights gained from these vulnerabilities could inform the development of defense strategies that ensure stable efficiency, particularly in resource-limited environments such as edge and mobile devices.

\noindent\textbf{Universal Segmentation among Multiple Domains.} 
{
The development of efficient SAM variants has been witnessed to achieve a great success, achieving significant improvement in runtime efficiency while demonstrating reasonable performance in image-level segmentation tasks.
However, most of these works only conduct accuracy evaluation on COCO and LVIS datasets, neglecting to assess efficient variants' universal segmentation capability across various domains.
To mitigate this gap, in our survey, we further evaluate models on two diverse, multi-domain benchmarks, Segmentation in the Wild (SGinW) and Unidentified Video Objects (UVO), to strengthen the evaluation towards SAM variants’ generalization performance in varied scenarios.
Results illustrated in Tab.~\ref{tab:sginw_transposed_rotated} and Tab.~\ref{tab:uvo} have shown that SAM variants may demonstrate disastrous performance when handling uncommon objects, for instance, several models falling short to segment Poles and Rail in SGinW benchmark, indicating efficient SAM variants' deficiency in rarely seen or unseen targets.
However, our experiments still do not reach the perfection, for not covering assessment of these variants' performance on some domain-specific tasks, such as segmentation in medical images or aerial images, which may further reveal their weakness in universal generalization.
Future work might focus on building up an efficient, almighty variant of SAM, boosting both the runtime efficiency and overall performance across a broader range of downstream tasks.
}

\noindent\textbf{Efficient Segmentation for Video and Multi-Modality Data.}
Video and multi-modality tasks, which deal with complex, dynamic environments, are rapidly gaining relevance across numerous real-world applications. While some initial efforts have extended SAM's applicability for video segmentation~\cite{ravi2024sam, jiaxing2025sam2, xu2025segment} and multi-modality tasks~\cite{xiao2024segment}, efficiency remains a pressing issue. Video data contains temporal redundancy, while multimodal data often exhibits correlations between modalities. Leveraging these inherent redundancies through techniques like temporal aggregation {\cite{sener2020temporal}} and cross-modal feature sharing {\cite{lu2020cross}} could significantly reduce computational costs. 
Future work could focus on optimizing SAM’s runtime complexity by utilizing spatiotemporal attention {\cite{diba2023spatio}}, efficient memory mechanisms {(which are now effective solutions for accelerating SAM 2~\cite{xiong2024efficientsam2, zhou2025edgetam})} for temporal data, 
and early-fusion techniques {\cite{hemkerhybrid}} to reduce the number of modality-specific computations. Developing frameworks that adapt dynamically to different levels of redundancy across frames or modalities can further drive computational efficiency in real-world applications.

\noindent\textbf{Efficient Variants of SAM 2}
Segment Anything Model 2, successor of Segment Anything Model, which not only achieves higher accuracy and efficiency on image segmentation task but also extends its powerful capability into video segmentation tasks, has been increasingly gaining attention from industry and research community.
However, similar to SAM's dilemma, SAM 2 also suffers the inefficiency issue due to the large hierarchical image encoder, hindering its application in various resource-constrained, real-world scenarios.
Moreover, another main bottleneck of SAM 2 lies in the newly introduced memory mechanism, more specifically, the memory attention module which attends feature map to memories, as memory features holds the same spatial size to image feature, resulting in long memory token sequence and heavy computational cost in cross-attention.
Several works of efficient SAM 2 variants have emerged focusing on solving these two bottlenecks.
The recently proposed Q-SAM2~\cite{farronato2025q} has introduced calibration in SAM 2's image encoder for backbone compression, while Surgical SAM 2~\cite{liu2024surgical} has proposed an frame pruning strategy for memory efficiency.
More variants like EfficientTAM~\cite{xiong2024efficientsam2}, EdgeTAM~\cite{zhou2025edgetam}, EffcientSAM2~\cite{zhang2026efficient} and TinySAM2~\cite{ding2026tinysam} take a two-pronged approach, compressing image encoder and improving memory mechanism simultaneously, for better acceleration.
Moreover, SAM 2 has also been rewritten by Pytorch Team~\cite{SAM2fast} with tricks like batching prompts and reducing precision, and techniques like AOTInductor’s (AOTI) ahead-of-time compilation, achieving noticeable enhancement on latency.
Compared to SAM, there are much less works on efficient variants for SAM 2 up till now, posing it a challenging task.
Techniques and strategies proved to be effective in accelerating SAM, have not been fully investigated on SAM 2, which might be the potential direction for future research.

\noindent\textbf{Efficient Variants of SAM 3}
For the newly published SAM 3~\cite{carion2025sam3segmentconcepts}, which generalized from its predecessor by combining a DETR-based detector with the SAM 2 style tracker, suffers similar efficiency bottlenecks as SAM 2 does. 
And the recently proposed method, EfficientSAM3~\cite{zeng2025efficientsam3}, has targeted at SAM3's inefficiency in its shared vision encoder and dense memory tracker. Two stages of distillation are leveraged to progressively lighten SAM 3, with a final stage of end-to-end fine-tuning for joint model refinement. 
The concurrent variant DART~\cite{turkcan2026detect} further accelerates SAM 3's multi-class inference, by leveraging backbone feature sharing, multi-class decoding, detection-only inference, and TensorRT deployment. 
Quantization-based techniques have also been applied to SAM3 in the recent work, Mix-QSAM3~\cite{ranjan2026mix}, which achieves great balance between accuracy and efficiency with a mixed-precision PTQ framework.

\section{Evaluation}
\label{evaluation}
In this section, we systematically compare the efficiency and accuracy of the previously described SAM variants. With reference to the experiments conducted in those works, we choose the tasks that most of the works have carried out and evaluate them on their commonly used datasets with corresponding metrics. Unless otherwise stated, our evaluations are conducted on a single \textit{24GB RTX 3090 GPU} with a 14 vCPU \textit{Intel(R) Xeon(R) Gold 6330 Processor @ 2.00GHz}. For fair comparisons, we ensure that all models are tested under the same environments and that all tested images are retained at their original resolutions. More details are provided in the following subsections: Section \ref{sec: 4.1} introduces the datasets and metrics used for evaluation; Sections \ref{sec: 4.2} and \ref{sec: 4.3} report the quantitative results of efficiency and accuracy, respectively.

\subsection{Datasets and Metrics}
\label{sec: 4.1}
{We utilize two fundamental datasets, COCO 2017~\cite{lin2014microsoft} and LVIS v1~\cite{gupta2019lvis}, and two more challenging benchmarks, SGinW\footnote{https://eval.ai/web/challenges/challenge-page/1931/overview} and UVO v1.0~\cite{wang2021unidentified}, for comprehensive evaluation. }
COCO is a large-scale dataset for object detection, segmentation, and captioning, containing 330K images and 1.5 million object instances across 80 object categories. LVIS is tailored for large vocabulary instance segmentation, featuring over 2 million high-quality segmentation masks across more than 1200 categories in 164K images. For our evaluation, we use the validation sets of both datasets, which include 36,781 instances in 5,000 images for COCO and 244,707 instances in 19,809 images for LVIS.

Segmentation in the Wild (SGinW) benchmark consists of 25 different segmentation datasets, {spanning a wide range of real-world domains, from common objects like fruits and animals to rare concepts like rail and brain tumor\cite{zou2023generalized}.} 
Unidentified Video Objects (UVO) is a challenging benchmark for dense, open-world segmentation in videos, {with numerous "unseen" objects not belonging to any of the COCO categories or the LVIS categories.}
It is worth noting that, for UVO benchmark, we only leverage the FrameSet for image-level evaluation, which includes 8124 frames sampled from 2708 videos with 65588 annotated instances.

To evaluate efficiency, we first test several soft indicators like \(\#Params\), \(FLOPs\), and \(Memory\  Usage\). 
And we further calculate the Efficient Error Rate (\(EER\)) for a more comprehensive evaluation, as outlined in \cite{papa2024survey},   
{to measure how much the variants are more efficient in comparison with the vanilla SAM-H. Specifically, \(EER\) is calculated by averaging the ratios of a set of efficient metrics between evaluated models and the reference model. The definition of \(EER\) is illustrated as follows, }
\begin{align} 
EER = \frac{1}{N}\sum_{i=1}^{N}(\frac{M_i}{R_i})
\label{eq:eer}
\end{align}
where \(N\) is the number of metrics and \(M_i\), \(R_i\) refer to the i-th metric of the tested model and the reference one, respectively. 
{
Consequently, tested models with higher efficiency will achieve lower \(EER\).  
In our evaluation, we select \{\(\#Params\), \(FLOPs\)\} as metrics and set the reference model to be SAM-H.}
In addition to these metrics, we also report the runtime and throughput of these models.
For accuracy evaluation, we use mean intersection over union (\(mIoU\)) for the SegAny task and mean average precision (\(AP\)) for instance segmentation.

\begin{table}[t]
    \centering
    \small
    \caption{Quantitative results of \(\#\textbf{Params (M)}\), \(\textbf{FLOPs (G)}\) and \(\textbf{EER}\). \(EER\) denotes the Efficient Error Rate proposed in \cite{papa2024survey} and is calculated by Eq. (\ref{eq:eer}), {which averaging the ratios of \{\#Params, FLOPs\} between variants and the vanilla SAM-H. The lower value of \(EER\) represent higher efficiency of the tested variant.}}
    \begin{tabular}{l|cc|c}
            \toprule
            {Model} & {Params} & \footnotesize{FLOPs} & \footnotesize{EER} \\ 
            \hline
            SAM-H  & 641.09 & 5490  & 100\% \\ 
            SAM-L  & 312.34 & 2640  & 48.40\% \\ 
            SAM-B  & 93.74 & 746.4  & 14.11\% \\ 
            SAM2-B+  & 80.83 & 533.87  & 11.17\% \\ 
            FastSAM\textsuperscript{*}  & 68 & 887.6 & 13.39\% \\ 
            MobileSAM  & 10.13 & 76.39 & 1.49\% \\ 
            \textbf{EdgeSAM}  & \textbf{9.58} & \textbf{44.02} & \textbf{1.15\%} \\ 
            EfficientSAM-Ti  & 10.22 & 53.64 & 2.57\% \\ 
            EfficientSAM-S  & 26.41 & 371.68 & 5.44\% \\ 
            RepVit-SAM  & 27.22 & 231.65 & 4.23\% \\ 
            Efficient\footnotesize{ViT-SAM-L0}  & 34.79 & 154.24 & 4.12\% \\ 
            Efficient\footnotesize{ViT-SAM-XL1}  & 203.35 & 523.07  & 20.62\% \\ 
            SlimSAM-50  & 28.01 & 196.21 & 3.97\% \\ 
            SlimSAM-77  & 9.85 & 47.66 & 1.20\% \\ 
            TinySAM  & 10.13 & 76.39 & 1.49\% \\ 
            \botrule
    \end{tabular}
    \footnotetext[*]{The results of FastSAM \cite{zhao2023fast} are collected from \cite{zhou2023edgesam} and others are evaluated under our framework.}
    \label{tab: soft}
\end{table}

\subsection{Efficiency Comparison}
\label{sec: 4.2}
We first report the efficiency results of SAM and its variants.
With the picture \textit{groceries.jpg}\footnote{https://github.com/facebookresearch/segment-anything/blob/main/notebooks/images/groceries.jpg} used in SAM's official example as input, we utilize a bounding box as prompt and evaluate models' \(\#Params\) and \(FLOPs\) by leveraging the tool \textit{calflops}\footnote{https://github.com/MrYxJ/calculate-flops.pytorch}. We then calculate the \(EER\) for further comparison. The results are illustrated in Tab. \ref{tab: soft}. 
Among the efficient variants, we observe that EdgeSAM has the lowest number of parameters, \(FLOPs\) and consequently the \(EER\), while EfficientViT-SAM-XL1 takes the highest numbers whose \(EER\) is even 6.51\% more than SAM-B. When compared to the heaviest SAM-H, all variants present obvious reduction in both model size and calculation.

\begin{figure}[!t]
    \centering
    \includegraphics[width=1\linewidth]{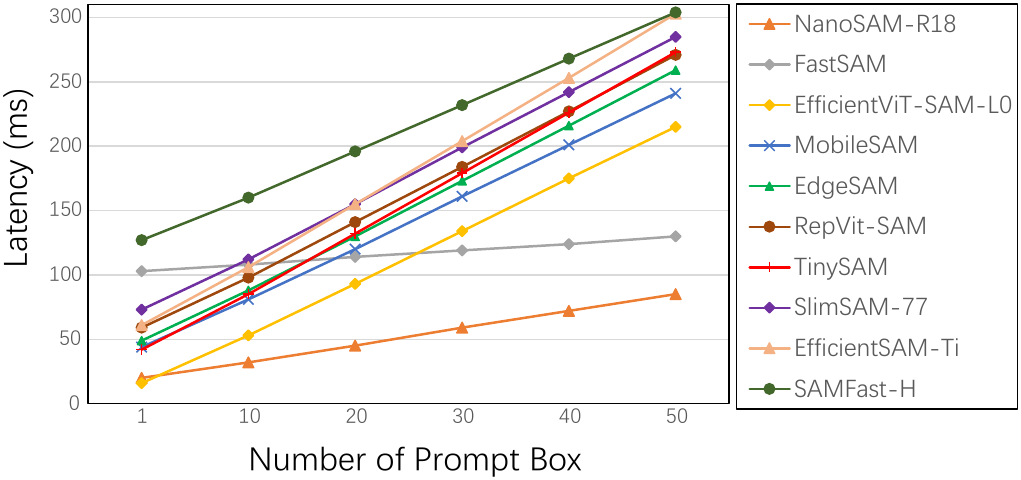}
    \caption{Latency results on GPU with increasing numbers of box prompts.}
    \label{fig:runtime-curve}
\end{figure}

We have also measured the models' inference time in both SegAny and SegEvery modes, using images from the COCO validation set as evaluation data. 
For the SegAny task, each image is prompted with 50 fixed bounding boxes. We record the cumulative time for every 10 boxes and illustrate the results through a curve chart (shown in Fig. \ref{fig:runtime-curve}). Based on this, we calculate the average time required to process one image with one box prompt and report it as the inference latency for the SegAny task. We conduct this evaluation on CPU-only environment, a series of single-GPU environments, and the edge device \(Jetson\ Nano\ 4GB\). We simultaneously report the approximation of GPU memory usage when running these models. 
Additionally, we test the throughput of each variant of conducting SegAny task on the COCO validation set, using ground truth bounding boxes as prompts.
The results are summarized in Tab. \ref{segAny efficiency}.
Our findings reveal that EfficientViT-SAM-L0 achieves the shortest inference time for the SegAny task, offering nearly 30x acceleration on GPU and almost 50x acceleration on CPU compared to the heaviest model, SAM-H. 
EdgeSAM also shows impressive performance with low latencies of 259 ms on CPU and 713 ms on the edge device, while NanoSAM achieves the lowest runtime on \(Jetson Nano\) and comes close to the best results on all GPU environments.
In the throughput test, NanoSAM leads with 27.9 images processed per second. Two other variants, EfficientSAM-Ti and EfficientViT-SAM-L0, also demonstrate strong throughput, each exceeding 20 images per second.

For the SegEvery task, we report the mean time required to generate all masks for an image using different point grid sizes (16*16, 32*32, 64*64) or specialized sampling strategies (detailed in Section \ref{sec: 3.2}). The results are presented in Tab. \ref{segEvery efficiency}. With the default 32*32 grid, SAMfast-H demonstrates the highest efficiency, with a latency of 848 ms—more than twice as fast as SAM-H. EfficientViT-SAM-L0 performs best with the 16*16 and 64*64 grids, achieving latencies of 258 ms and 3938 ms, respectively.
Interestingly, we observe that EfficientSAM-S is slower than SAM-H when using lower grid densities, showing latencies of 1100 ms for the 16*16 grid and 2290 ms for the 32*32 grid. Models employing alternative sampling strategies demonstrate significant improvements in efficiency, particularly FastSAM, which records a latency of 106 ms, and MobileSAMv2, with a latency of 173 ms.

\begin{table*}[ht]
    \centering
    \small
    \caption{Quantitative results of \textbf{inference latency (ms)} and of SegAny task, \textbf{throughput (img/s)} on COCO validation dataset, and the approximation of \textbf{GPU memory usage (GB)}. More details are described in Section~\ref{sec: 4.2}. For the latency comparison, we report the results on a CPU-only environment, various GPU-based environments (1080Ti/11GB, 2080Ti/11GB, 3090/24GB, and A800/80GB), and the edge device (Jetson Nano 4GB). The throughput is measured with a single 3090 GPU.
    }
    \begin{tabular*}{\textwidth}{@{\extracolsep{\fill}} l|c|cccccc|c @{}}
        \toprule
        ~ & ~ & \multicolumn{6}{c|}{Latency} & ~ \\
        Model & Memory & CPU & Nano & 1080Ti & 2080Ti & 3090 & A800 & Throughput\\       
        \hline
        SAM-H  & 7.46  & 9470 & - & 1039 & 711 & 461 & 431 & 2.04  \\ 
        SAM-L  & 6.03  & 6800 & - & 603 & 400 & 270 & 248 & 3.33  \\ 
        SAM-B  & 4.39  & 3294 & - & 270 & 170 & 116 & 103 & 6.78  \\ 
        SAMfast-H  & 2.13  & - & - & - & 939 & 127 & 57 & 6.36  \\ 
        SAM2-B+  & 1.38 & 1221 & - & 197 & 110 & 85 & 61 & 13.82  \\ 
        FastSAM  & 4.96  & 850 & - & 183 & 142 & 103 & 53 & 8.42  \\ 
        MobileSAM  & 1.81  & 424 & 1328 & 60 & 38 & 30 & 25 & 16.79  \\ 
        EdgeSAM  & 1.72  & 259 & 713 & 40 & 32 & 24 & 24 & 17.56  \\ 
        EfficientSAM-Ti  & 4.46 & 1159 & - & 92 & 51 & 40  & 30 & 20.04  \\ 
        EfficientSAM-S  & 7.53  & 2605 & - & 188 & 100 & 73 & 60 & 19.7  \\ 
        RepVit-SAM  & 1.87  & 1013 & 2685 & 104 & 69 & 44 & 38 & 13.49  \\ 
        EfficientViT-SAM-L0  & 1.73  & \textbf{194} & 651 & \textbf{30} & \textbf{23} & \textbf{16} & \textbf{15} & 20.92  \\ 
        EfficientViT-SAM-XL1  & 2.57  & 1334 & 3779 & 116 & 82 & 52 & 36 & 12.28  \\ 
        SlimSAM-50  & 4.49  & 2312 & - & 192 & 116 & 76 & 68 & 9.13  \\ 
        SlimSAM-77  & 4.34  & 2230 & - & 161 & 94 & 65 & 55 & 10.28  \\ 
        TinySAM  & 2.2  & 422 & 1322 & 60 & 39 & 29 & 25 & 16.53  \\ 
        Q-TinySAM  & 2.34  & 697 & 2217 & 116 & 78 & 58  & 48 & 7.03  \\ 
        NanoSAM  & \textbf{0.94}  & - & \textbf{345} & 39 & 26& 20 & 16 & \textbf{27.9 } \\ 
        \bottomrule
    \end{tabular*}
    \label{segAny efficiency}
\end{table*}

\begin{table}[ht]
    \centering
    \small
    \caption{Quantitative results of \textbf{inference latency (ms)} of SegEvery task with grid sampling in different scales or other efficient strategies. More details are reviewed in Section \ref{sec: 3.2}.}
    \begin{tabular}{l|cccc}
        \toprule
         ~ & \multicolumn{3}{c}{Grid Sampling} & Other\\
         Model & $16^2$ & $32^2$ & $64^2$ &  \\ 
        \hline
        SAM-H  & 824 & 2269 & 8103 & - \\ 
        SAM-L  & 665 & 2053 & 7518 & - \\ 
        SAM-B  & 422 & 1399 & 5417 & - \\ 
        SAMfast-H  & 315 & \textbf{848} & OOM\textsuperscript{*} & - \\ 
        SAM2-B+  & 442 & 1610 & 6356 & - \\ 
        FastSAM  & - & - & - & \textbf{106}  \\ 
        MobileSAM  & 347 & 1313 & 5022 & ~ \\ 
        MobileSAMv2 & - & - & - & 173  \\ 
        EdgeSAM  & 404 & 1553 & 5877 & - \\ 
        EfficientSAM-Ti  & 753 & 1833 & 5670 & - \\ 
        EfficientSAM-S  & 1100 & 2290 & 6210 & - \\ 
        RepVit-SAM  & 512 & 1890 & 7437 & - \\ 
        \footnotesize{EfficientViT-SAM-L0}  & \textbf{258} & 1023 & \textbf{3938} & - \\ 
        \footnotesize{EfficientViT-SAM-XL1}  & 432 & 1546 & 6094 & - \\ 
        SlimSAM-50  & 400 & 1374 & 5213 & - \\ 
        SlimSAM-77  & 385 & 1342 & 5148 & - \\ 
        TinySAM  & - & - & - & 776  \\ 
        Q-TinySAM  & - & - & - & 1318  \\ 
        \toprule
    \end{tabular}
    \footnotetext[*]{OOM denotes out of memory. }
    \label{segEvery efficiency}
\end{table}

\subsection{Accuracy Comparison}
\label{sec: 4.3}
In this subsection, we report the accuracy results of SAM and its variants on SegAny task (with point/box prompts) and instance segmentation task,  respectively. We follow the evaluation framework in \cite{zhang2024efficientvit} and adapt it by introducing other evaluation modules to conduct a unified evaluation on variants.

\begin{table*}[ht]
    \centering
    \small
    \caption{Quantitative results of the accuracy of SegAny task (\textbf{mIoU}) on COCO and LVIS with points and boxes as prompts. 
    For evaluation with points prompts, we select the center point of the ground truth bounding box (\(pt_1\)), and one or three randomly sampled points from ground truth masks (1*\(pt_2\), 3*\(pt_2\)).
    For evaluation with boxes prompts, we adopt the ground truth bounding box (\(box_1\)), and the  tightest bounding box calculated by the ground truth mask (\(box_2\)).}
    \begin{tabular*}{\textwidth}{@{\extracolsep{\fill}}l|ccc|cc|ccc|cc@{} }
    \toprule
         ~ & \multicolumn{5}{c|}{COCO} & \multicolumn{5}{c}{LVIS}\\ 
        Model & \(pt_1\) & 1*\(pt_2\) & 3*\(pt_2\) &  \(box_1\) & \(box_2\) & \(pt_1\) & 1*\(pt_2\) & 3*\(pt_2\) & \(box_1\) & \(box_2\) \\ 
        \hline
        SAM-H  & 53.6  & 55.6  & 67.5 & 77.4  & 77.8 & 60  & 59.8  & 65.0 & 78.0  & 79.1 \\ 
        SAMfast-H  & 52.9  & 54.5 & 67.4 & 77.3  & 77.7 & \textbf{53.6}  & \textbf{57.2} & 60.6  & 77.0  & 78.1\\ 
        SAM2-B+  & \textbf{54.3}  & 54.9 & 68.5 & 75.7  & 76.6 & 53.1  & 55.9 & 62.6  & 70.4  & 72.4 \\ 
        FastSAM  & 46.7  & 47.3 & 48.7 & 60.5  & 60.6 & 28.4  & 29.8  & 29.5 & 50.4  & 50.6 \\ 
        MobileSAM  & 48.6  & 49.2 & 59.1 & 73.4  & 73.9 & 45.6  & 47.2 & 50.0 & 74.4  & 74.8\\ 
        EdgeSAM  & 47.2  & 46.8 & 61.6 & 75.9  & 76.2 & 45.2  & 46.5 & 54.4 & 74.4  & 74.8 \\ 
        EfficientSAM-Ti  & 45.7  & 44.7 & 66.6 & 74.7  & 75.3 & 51.5  & 53.0 & \textbf{67.5} & 74.4  & 76.0\\ 
        EfficientSAM-S  & 52.9  & 52.5 & 68.6  & 76.1  & 76.6 & 52.5  & 54.4 & 64.7 &  74.7  & 76.0\\ 
        RepViT-SAM  & 50.6  & 51.8 & 63.1 & 75.1  & 75.7 & 51.0  & 51.6 & 54.6 & 73.6  & 74.8\\ 
        EfficientViT-SAM-L0  & 50.1  & 50.7 & 67.2 & 78.5  & 78.2 & 46.4  & 48.3 & 63.9  & 78.0  & 77.9 \\ 
        EfficientViT-SAM-XL1  & \textbf{54.3}  & \textbf{55.7} & \textbf{70.2}  & \textbf{79.9}  & \textbf{79.8} & 53.5  & 54.4 & 65.7 & \textbf{79.9} & \textbf{79.7} \\ 
        SlimSAM-50  & 48.5  & 48.5 & 63.5 & 75.5  & 76.3 & 51.7  & 53.4 & 58.6 & 74.3  & 76.0\\ 
        SlimSAM-77 & 47.2  & 47.2 &  61.1 & 74.4 & 75.2 & 48.6 & 50.4 & 55.3 & 72.7  &  74.5 \\ 
        TinySAM  & 42.4  & 42.1 & 65.1  & 73.7  & 73.9  & 50.0  & 51.0  & 60.9  & 73.6  & 74.1  \\ 
        NanoSAM  & 44.2  & 42.4 & 56.3 & 69.7  & 70.2 & 42.7  & 41.4 & 48.3  & 65.1  & 66.5 \\ 
    \bottomrule
    \end{tabular*}
    \label{point-miou}
\end{table*}

\begin{table*}[ht]
    \small
    \centering
    \caption{Quantitative results of instance segmentation on COCO and LVIS with ViTDet \cite{li2022exploring} as object detector.}
     \begin{tabular*}{\textwidth}{@{\extracolsep{\fill}} l|cccc|cccc @{}}
     \toprule
         & \multicolumn{4}{c|}{COCO} & \multicolumn{4}{c}{LVIS}  \\
        Model& \(\rm AP\) & \(\rm AP^S\) & \(\rm AP^M\) & \(\rm AP^L\) & \(\rm AP\) & \(\rm AP^S\) & \(\rm AP^M\) & \(\rm AP^L\) \\ 
        \hline
        SAM-H  & 46.5 & 30.8 & 51.0 & 61.7 & 44.7 & 32.5 & 57.6 & 65.5  \\ 
        SAMfast-H  & 46.4 & 30.4 & 50.8 & 61.9 & \textbf{44.5} & \textbf{32.3} & \textbf{57.5} & 65.7  \\ 
        SAM2-B+  & 44.8 & 27.1 & 49.9 & 62.5 & 42.3 & 30.1 & 57.0 & 65.3  \\ 
        FastSAM  & 37.9 & 23.9 & 43.4 & 50.0 & 34.5 & 24.6 & 46.2 & 50.8  \\ 
        MobileSAM  & 38.7 & 23.7 & 42.2 & 54.3 & 37.0 & 24.7 & 47.8 & 59.1  \\ 
        EdgeSAM  & 42.1 & 26.6 & 46.7 & 56.9 & 39.8 & 28.6 & 51.3 & 59.3  \\ 
        EfficientSAM-Ti  & 42.1 & 26.5 & 45.5 & 57.7 & 39.9 & 28.9 & 51.0 & 59.4  \\ 
        EfficientSAM-S  & 44.5 & 28.5 & 48.7 & 60.1 & 41.5 & 29.7 & 53.4 & 62.2  \\ 
        RepViT-SAM  & 43.3 & 27.0 & 47.4 & 59.6 & 40.2 & 28.0 & 52.3 & 61.4  \\ 
        EfficientViT-SAM-L0  & 45.7 & 28.2 & 49.5 & 63.4 & 41.8 & 28.8 & 53.4 & 64.7  \\ 
        EfficientViT-SAM-XL1  & \textbf{47.8} & \textbf{30.5} & \textbf{51.8} &\textbf{ 64.7} & 44.4 & 31.6 & 57.0 & \textbf{66.4}  \\ 
        SlimSAM-50  & 42.8 & 27.2 & 46.8 & 58.4 & 40.1 & 28.5 & 51.8 & 60.5  \\ 
        SlimSAM-77  & 41.3 & 25.7 & 44.9 & 57.5 & 38.3 & 26.8 & 49.7 & 59  \\ 
        TinySAM  & 42.3 & 26.3 & 45.8 & 58.8 & 38.9 & 27.0 & 50.3 & 60.2  \\ 
        Q-TinySAM  & 41.4 & 25.6 & 45.1 & 57.9 & 38.5 & 26.6 & 49.8 & 59.8  \\ 
        NanoSAM  & 35.9 & 20.0 & 40.3 & 52.4 & 31.6 & 19.5 & 42.7 & 53.7  \\ 
    \bottomrule
    \end{tabular*}
    \label{box-vitdet}
\end{table*}


To evaluate on the SegAny task, we adopt two types of points as prompts: 1) the center point of the ground truth bounding box, and 2) random points uniformly sampled from ground truth masks, following the setting in \cite{xiong2024efficientsam}. We evaluate the variants on COCO and LVIS, and the mean intersection over union (mIoU) is reported in Tab.~\ref{point-miou}.
When prompted with the center point, SAM2-B+ and EfficientViT-SAM-XL1 achieve the highest mIoU of 54.3\% on COCO dataset, surpassing SAM-H with 53.6\% mIoU, while on the LVIS dataset, SAMfast-H, with mIoU of 53.6\%, exhibits the best performance among variants.
Under the setting of random point prompts, it is demonstrated that EfficientViT-SAM-XL1 performs better than SAM-H on both datasets when prompted with 3 points, with the increase of 2.7\% and 0.7\%, respectively. From the perspective of datasets, we observe that the results on LVIS are generally lower than those on COCO. 

Moreover, we also evaluate the accuracy on SegAny task with two types of box prompts: 1) The ground truth bounding box and 2) The tightest bounding box corresponding to the ground truth mask, instructed by the experiments in \cite{xiong2024efficientsam, zhang2024efficientvit}. We reported the results of mIoU on COCO and LVIS which are illustrated in Tab. \ref{point-miou}. We observe that EfficientViT-SAM-XL1 demonstrate the highest accuracy in every box-prompted setting, which exceeds SAM-H by 2.5\%, 2.0\%, 1.9\% and 0.6\%, respectively. SAMfast-H and EfficientViT-SAM-L0 also present performance close to SAM-H in the segmentation tasks with box prompts.

\begin{table*}[t]
    \small
    \centering
    \setlength{\tabcolsep}{3pt}
    \caption{Quantitative results of instance segmentation on COCO with YOLOv8 \cite{yolov8} or GrounddingDINO \cite{liu2023grounding} as object detector.}
     \begin{tabular*}{\textwidth}{@{\extracolsep{\fill}} l|cccc|cccc @{}}
    \toprule
         & \multicolumn{4}{c|}{YOLOv8} & \multicolumn{4}{c}{GroundingDINO}  \\
        Model& \(\rm AP\) & \(\rm AP^S\) & \(\rm AP^M\) & \(\rm AP^L\) & \(\rm AP\) & \(\rm AP^S\) & \(\rm AP^M\) & \(\rm AP^L\) \\ 
    \hline
        SAM-H  & 43.8 & 26.1 & 48.1 & 60.4  & 46.9 & 31.5 & 51.8 & 64.4  \\ 
        SAMfast-H  & 43.6 & \textbf{26.0} & 47.9 & 60.2  & 46.8 & \textbf{31.5} & 51.6 & 64.2  \\ 
        SAM2-B+  & 42.4 & 24.1 & 47.4 & 60.5  & 45.1 & 28.2 & 50.8 & 64.6  \\ 
        FastSAM  & 34.2 & 19.7 & 40.1 & 46.6  & 35.7 & 22.1 & 42.1 & 48.2  \\ 
        MobileSAM  & 39.0 & 21.7 & 42.8 & 56.5  & 41.4 & 25.7 & 45.6 & 60.5  \\ 
        EdgeSAM  & 39.5 & 22.6 & 43.7 & 56.0  & 42.5 & 27.4 & 46.9 & 59.5  \\ 
        EfficientSAM-Ti  & 40.0 & 23.5 & 43.6 & 56.5 & 42.9 & 27.6 & 46.9 & 60.6  \\ 
        EfficientSAM-S  & 42.1 & 24.9 & 46.6 & 58.4 & 45 & 29.7 & 49.6 & 62.2  \\ 
        RepViT-SAM  & 41.0 & 23.8 & 45.1 & 57.9  & 43.8 & 28.3 & 48.2 & 61.6  \\ 
        EfficientViT-SAM-L0  & 42.7 & 24.1 & 46.8 & 61.2  & 46.0 & 29.2 & 50.3 & 65.7  \\ 
        EfficientViT-SAM-XL1  & \textbf{44.7} & \textbf{26.0} & \textbf{48.9} & \textbf{62.9}  & \textbf{48.2} & \textbf{31.5} & \textbf{52.6} & \textbf{67.3}  \\ 
        SlimSAM-50  & 40.6 & 23.9 & 44.7 & 57.0  & 43.5 & 28.7 & 47.8 & 60.8  \\ 
        SlimSAM-77  & 39.3 & 22.5 & 43.1 & 55.9  & 42.1 & 27.1 & 46.0 & 59.6  \\ 
        TinySAM  & 39.7 & 22.8 & 43.5 & 56.7  & 42.4 & 27.4 & 46.5 & 60.6  \\ 
        NanoSAM  & 34.3 & 18.0 & 38.4 & 50.9  & 36.3 & 21.2 & 41.2 & 54.5  \\ 
    \bottomrule
    \end{tabular*}
    \label{box-yolov8/groundingdino}
\end{table*}

\begin{table*}[t]
    \small
    \centering
    \setlength{\tabcolsep}{3pt}
    \caption{Quantitative results of instance segmentation on COCO with Detic \cite{zhou2022detecting} or H-Deformable-DETR \cite{jia2023detrs} as the object detector.}
    \begin{tabular*}{\textwidth}{@{\extracolsep{\fill}} l|cccc|cccc @{}}
    \toprule
         & \multicolumn{4}{c|}{Detic} & \multicolumn{4}{c}{H-Deformable-DETR}  \\
        Model& \(\rm AP\) & \(\rm AP^S\) & \(\rm AP^M\) & \(\rm AP^L\) & \(\rm AP\) & \(\rm AP^S\) & \(\rm AP^M\) & \(\rm AP^L\) \\ 
    \hline
        SAM-H  & 39.2 & 26.7 & 44.4 & 51.3 & 46.8 & 31.8 & 51.0 & 63.6  \\ 
        FastSAM  & 31.1 & 19.4 & 38.6 & 40.9 & 35.3 & 21.6 & 41.2 & 48.0  \\ 
        MobileSAM  & 34.9 & 22.7 & 39.8 & 47.6 & 42.7 & 27.0 & 46.5 & 61.1  \\ 
        EdgeSAM  & 35.2 & 23.5 & 40.3 & 46.6 & 32.1 & 21.6 & 44.3 & 59.4  \\ 
        EfficientSAM-Ti  & 35.4 & 24.3 & 40.2 & 47.0 & 43.6 & 27.3 & 47.3 & 61.1 \\ 
        EfficientSAM-S  & 37.5 & 25.1 & 42.7 & 49.5 & 45.4 & 29.8 & 49.7 & 62.7  \\ 
        RepViT-SAM  & 36.5 & 24.7 & 41.7 & 48.2 & 44.4 & 29.1 & 48.6 & 61.4  \\ 
        EfficientViT-SAM-L0  & 37.9 & 24.1 & 42.7 & 51.6 & 45.8 & 29.2 & 50 & 64.6 \\ 
        EfficientViT-SAM-XL1  & \textbf{39.5} & \textbf{26.3} & \textbf{44.6} & \textbf{52.2}  & \textbf{47.9} & \textbf{31.4} & \textbf{52.1} & \textbf{64.6} \\ 
        SlimSAM-50  & 34.3 & 24.9 & 40.5 & 43.1 & 42.2 & 29.3 & 46.8 & 56.5  \\ 
        SlimSAM-77  &  34.9 & 22.7 & 39.6 & 47.0 & 26.8 & 21.6 & 43.9 & 59.0 \\ 
        TinySAM  & 35.2 & 23.2 & 39.9 & 47.5 & 43.1 & 27.4 & 46.6 & 61.2  \\ 
        NanoSAM & 25.1 & 13.2 & 31.9 & 37.6  & - & - & - & -\\ 
    \bottomrule
    \end{tabular*}
    \label{box-Detic/H-Deformable-DETR}
\end{table*}

For instance segmentation tasks, we adopt the ViTDet \cite{li2022exploring}, YOLOv8 \cite{yolov8}, GrounddingDINO \cite{liu2023grounding}, Detic \cite{zhou2022detecting} and H-Deformable-DETR \cite{jia2023detrs} with Swin-L \cite{liu2021swin} as object detectors to generate bounding boxes of latent objects, with reference to the experiments conducted in \cite{wang2023repvit, zhou2023edgesam, zhang2024efficientvit}. We evaluate the average precision (AP) of all objects, as well as AP of small, medium, and large objects. The results are reported in Tab. \ref{box-vitdet}, \ref{box-yolov8/groundingdino} and \ref{box-Detic/H-Deformable-DETR}. Similar to the previous results, we find that on COCO dataset EfficientViT-SAM-XL1 always presents the highest AP with any of the detectors. Under the setting of equipping ViTDet as detector and testing on LVIS dataset, SAMfast-H surpass all other variants with AP of 44.5\%.

{
To further evaluate variants' segmentation ability in more challenging scenarios, we conduct instance segmentation on SGinW and UVO benchmark, equipped with GrounddingDINO for box prompts generation by following~\cite{ke2024segment}.
Results are illustrated in Tab.~\ref{tab:sginw_transposed_rotated} and Tab.~\ref{tab:uvo}, respectively. 
On SGinW benchmark, we observe that the rewritten SAMfast-H achieves the highest mean AP over 25 datasets, with only 0.1\% decrease compared to SAM-H, followed by EfficientViT-SAM-XL1 and EfficientSAM-S with 0.4\% and 0.8\% more drop respectively. 
From the perspective of dataset, the segmentation results on House-Parts, Brain-Tumor and Rail are significantly lower than others, indicating them more challenging datasets. 
On UVO benchmark, we notice that EfficientViT-SAM-XL1 demonstrates the best performance with AP of 31\%, exceeding SAM-H by 1.1\%, while SAM2.1-B+ also surpasses original SAM by 1.0\%.
}

\begin{table*}[ht]
\centering
\small
\setlength{\tabcolsep}{3pt}
\caption{Quantitative results of zero-shot instance segmentation on SGinW benchmark with GroundingDINO as the object detector.We report variants' Average Precision (AP) on each dataset and mean AP over all 25 datasets.}
\label{tab:sginw_transposed_rotated}
\begin{tabular}{@{} l | *{16}{c} @{}}
\toprule
 {Dataset} & 
{\footnotesize \rotatebox{90}{ {SAM-H}}} & 
{\footnotesize \rotatebox{90}{ {SAMfast-H}}} & 
{\footnotesize \rotatebox{90}{ {sam2-B+}}} & 
{\footnotesize \rotatebox{90}{ {FastSAM}}} & 
{\footnotesize \rotatebox{90}{ {MobileSAM}}} & 
{\footnotesize \rotatebox{90}{ {EdgeSAM}}} & 
{\footnotesize \rotatebox{90}{ {EfficientSAM-Ti}}} & 
{\footnotesize \rotatebox{90}{ {EfficientSAM-S}}} & 
{\footnotesize \rotatebox{90}{ {RepViT-SAM}}} & 
{\footnotesize \rotatebox{90}{ {SlimSAM50}}} & 
{\footnotesize \rotatebox{90}{ {SlimSAM77}}} & 
{\footnotesize \rotatebox{90}{ {TinySAM}}} & 
{\footnotesize \rotatebox{90}{ {Q-TinySAM}}} & 
{\footnotesize \rotatebox{90}{ {EfficientViT-SAM-L0}}} & 
{\footnotesize \rotatebox{90}{ {EfficientViT-SAM-XL1}}} & 
{\footnotesize \rotatebox{90}{ {NanoSAM}}}  \\
\hline
 {Airplane-Parts}    & 35.3  & 35.2   & \textbf{36.6}   & 23.1      & 31.1       & 31.6      & 32.5            & 35.1           & 29.6      & 29.1       & 26.1       & 28.6     & 33.0       & 35.2               & 34.6               & 30.5      \\
 {Bottles}           & 65.7  & 65.2   & 66.5   & 59.6      & 64.1       & 64.9      & 63.4            & 64.8           & 65.1      & 63.8       & 64.0       & 64.6     & 63.1       & 66.5               & \textbf{67.1}               & 63.7      \\
 {Brain-Tumor}       & 11.7  & 11.7   & 11.2   & 2.1       & \textbf{12.2}       & 11.3      & 12.0            & 12.0           & 11.6      & 11.2       & 11.6       & 9.9      & 11.8       & 11.5               & 12.1               & 11.9      \\
 {Chicken}           & 85.9  & \textbf{86.8}   & 79.1   & 71.1      & 84.5       & 81.9      & 83.4            & 85.6           & 85.2      & 83.2       & 85.5       & 83.2     & 82.7       & 82.4               & 85.1               & 80.2      \\
 {Cows}              & 46.3  & 46.2   & \textbf{46.6}   & 17.0      & 42.7       & 40.1      & 43.3            & 45.3           & 42.2      & 43.0       & 41.5       & 38.2     & 41.4       & 43.5               & 45.3               & 40.5      \\
 {Electric-Shaver}   & 70.5  & 70.2   & \textbf{71.8}   & 49.4      & 66.3       & 65.2      & 66.2            & 68.5           & 64.2      & 58.1       & 63.9       & 59.9     & 65.6       & 69.8               & 70.6               & 64.2      \\
 {Elephants}         & 77.2  & \textbf{77.3}   & 76.2   & 61.3      & 70.5       & 68.4      & 70.7            & 74.1           & 70.0      & 68.6       & 70.1       & 64.3     & 68.4       & 72.2               & 74.8               & 68.1      \\
 {Fruits}            & 82.3  & \textbf{82.3}   & 80.7   & 81.5      & \textbf{82.3}       & 81.5      & 81.7            & 81.7           & 81.5      & 81.5       & 81.5       & 74.0     & 76.2       & 81.5               & 81.5               & \textbf{82.3}     \\
 {Garbage}           & 23.7  & 23.9   & \textbf{25.9}   & 10.1      & 22.7       & 20.5      & 19.7            & 23.3           & 21.9      & 17.8       & 19.2       & 16.9     & 21.8       & 23.5               & 23.2               & 20.7      \\
 {Ginger-Garlic}     & 45.8  & 45.8   & 44.9   & 28.3      & 43.6       & 40.2      & 44.6            & 45.7           & \textbf{47.0}      & 39.5       & 40.8       & 39.4     & 46.1       & 43.9               & 46.7               & 42.6      \\
 {Hand}              & 76.6  & 76.5   & 78.2   & \textbf{81.6}      & 73.2       & 71.0      & 69.3            & 76.7           & 73.0      & 71.9       & 74.6       & 66.9     & 70.8       & 74.1               & 75.3               & 70.7      \\
 {Hand-Metal}        & 81.4  & 81.1   & 78.8   & 70.6      & 80.9       & 78.5      & 76.4            & 80.0           & \textbf{81.3}      & 77.8       & 78.8       & 78.7     & 78.4       & 79.2               & 79.6               & 78.5      \\
 {HouseHold-Items}   & 60.1  & \textbf{60.1}   & \textbf{60.1}   & \textbf{60.1}      & 58.8       & 58.8      & 58.8            & \textbf{60.1}           & \textbf{60.1}      & \textbf{60.1}       & \textbf{60.1}       & 58.8     & 59.5       & \textbf{60.1}               & \textbf{60.1}               & 58.8      \\
 {House-Parts}       & 8.4   & \textbf{8.3}    & 8.0    & 6.6       & 8.1        & 8.1       & 8.0             & \textbf{8.3}            & \textbf{8.3}       & 8.1        & 8.0        & 7.8      & 7.9        & 8.2                & \textbf{8.3}                & 8.1       \\
 {Nutterfly-Squireel} & 70.0  & 69.4   & \textbf{80.8}   & 23.1      & 63.1       & 57.5      & 68.5            & 70.8           & 59.7      & 52.8       & 53.7       & 59.5     & 60.4       & 63.5               & 67.8               & 60.3      \\
 {Phones}            & 34.9  & 35.0   & 30.2   & 28.4      & 33.9       & 33.4      & 34.0            & \textbf{35.5}           & 34.5      & 33.8       & 35.3       & 34.2     & 34.5       & 34.3               & 34.8               & 31.4      \\
 {Poles}             & 23.3  & 23.3   & 15.6   & 5.1       & 16.6       & 15.9      & 20.0            & \textbf{27.7}           & 13.4      & 14.2       & 12.5       & 4.7      & 12.5       & 19.0               & 16.7               & 6.9       \\
 {Puppies}           & 50.1  & \textbf{50.1}   & \textbf{50.1}   & 28.4      & 45.0       & 44.6      & 46.7            & 47.8           & 45.0      & 46.0       & 46.8       & 38.6     & 44.2       & 45.8               & \textbf{50.1}               & 43.8      \\
 {Rail}              & 15.8  & 15.7   & 12.2   & 1.0       & 14.1       & 13.5      & 12.9            & 13.4           & 12.9      & 12.4       & 15.1       & \textbf{17.5}     & 15.4       & 17.5               & 16.7               & 13.1      \\
 {Salmon-Fillet}     & 36.2  & 36.3   & 19.7   & 24.5      & 29.9       & 31.7      & 12.4            & 21.8           & 25.2      & 27.8       & 28.0       & 18.3     & 27.0       & \textbf{39.2}               & 37.1               & 25.6      \\
 {Strawberry}        & 83.5  & 83.3   & 83.6   & 79.8      & 83.3       & 83.0      & 82.7            & 83.3           & 83.6      & 82.6       & 82.4       & \textbf{83.7}     & 82.6       & 83.4               & 83.6               & 83.5      \\
 {Tablets}           & 29.6  & \textbf{29.5}   & 27.1   & 29.0      & 28.7       & 28.4      & 29.0            & 29.3           & 28.9      & 28.7       & 28.7       & 28.6     & 28.5       & 28.8               & 28.8               & 28.7      \\
 {Toolkits}          & 20.9  & \textbf{20.9}   & 20.2   & 12.3      & 19.3       & 18.6      & 19.1            & 20.7           & 18.5      & 17.7       & 18.6       & 19.2     & 18.6       & 19.9               & 20.6               & 19.1      \\
 {Trash}             & 29.7  & \textbf{29.7}   & 27.7   & 27.3      & 28.8       & 28.3      & 28.8            & 29.1           & 28.9      & 28.6       & 28.7       & 27.5     & 28.2       & 28.7               & 29.3               & 28.6      \\
 {Watermelon}        & 64.9  & 64.5   & 63.7   & 44.4      & 63.5       & 66.2      & 64.6            & 66.1           & 62.7      & 62.1       & 63.7       & 62.8     & 63.4       & \textbf{67.3}               & 67.0               & 63.2      \\
\hline
 {Mean AP}           & 49.2  &  \textbf{49.1} & 47.8   & 37.0      & 46.7       & 45.7      & 45.9            & 48.3           & 46.2      & 44.8       & 45.6       & 43.4     & 45.7       & 48.0               & 48.7               & 45.0      \\
\bottomrule
\end{tabular}
\end{table*}

\begin{table}[t]
    \centering
    \small
    \caption{Quantitative results of zero-shot instance segmentation on UVO benchmark with GroundingDINO as the object detector. }
    \begin{tabular}{l|cccc}
        \toprule
         Model & \(\rm AP\) & \(\rm AP^S\) & \(\rm AP^M\) & \(\rm AP^L\) \\ 
        \hline
        SAM-H  & 29.9 &	10 & 20.8 & 44.9 \\ 
        SAMfast-H & 29.7 & \textbf{9.9} & 20.7 & 44.6\\ 
        SAM2-B+  & 30.9 &	9.4 & \textbf{21.3} & 47\\ 
        FastSAM  & 20.8 & 7 & 14.7 & 30.1  \\ 
        MobileSAM & 25.2 & 8.2 & 17.4 & 38\\ 
        EdgeSAM  & 24.9 & 8.6 & 17.9 & 36.4\\ 
        EfficientSAM-Ti  & 25.3 & 7.5 & 17.9 & 38.1\\ 
        EfficientSAM-S  & 28 & 9.5 & 19.3 & 41.9\\ 
        RepVit-SAM  & 26.2 & 8.9 & 18.2 & 39.1\\ 
        \footnotesize{EfficientViT-SAM-L0}  & 29.5 & 9.2 & 19.9 & 45.1\\ 
        \footnotesize{EfficientViT-SAM-XL1}  & \textbf{31} & 9.8 & 21.1 & \textbf{47.2} \\ 
        SlimSAM-50  & 26.2 & 9.2 & 18.6 & 38.6\\ 
        SlimSAM-77  & 25.1 & 8.7 & 17.8 & 37 \\ 
        TinySAM  & 22.9 & 8.1 & 16.5 & 33.4 \\ 
        Q-TinySAM  & 24.8 & 8 & 17.6 & 37 \\ 
        NanoSAM & 22.9 & 6.2 & 15.4 & 35.5 \\
        \toprule
    \end{tabular}
    \label{tab:uvo}
\end{table}

\begin{figure}[ht]
    \centering
    \includegraphics[width=1.05\linewidth]{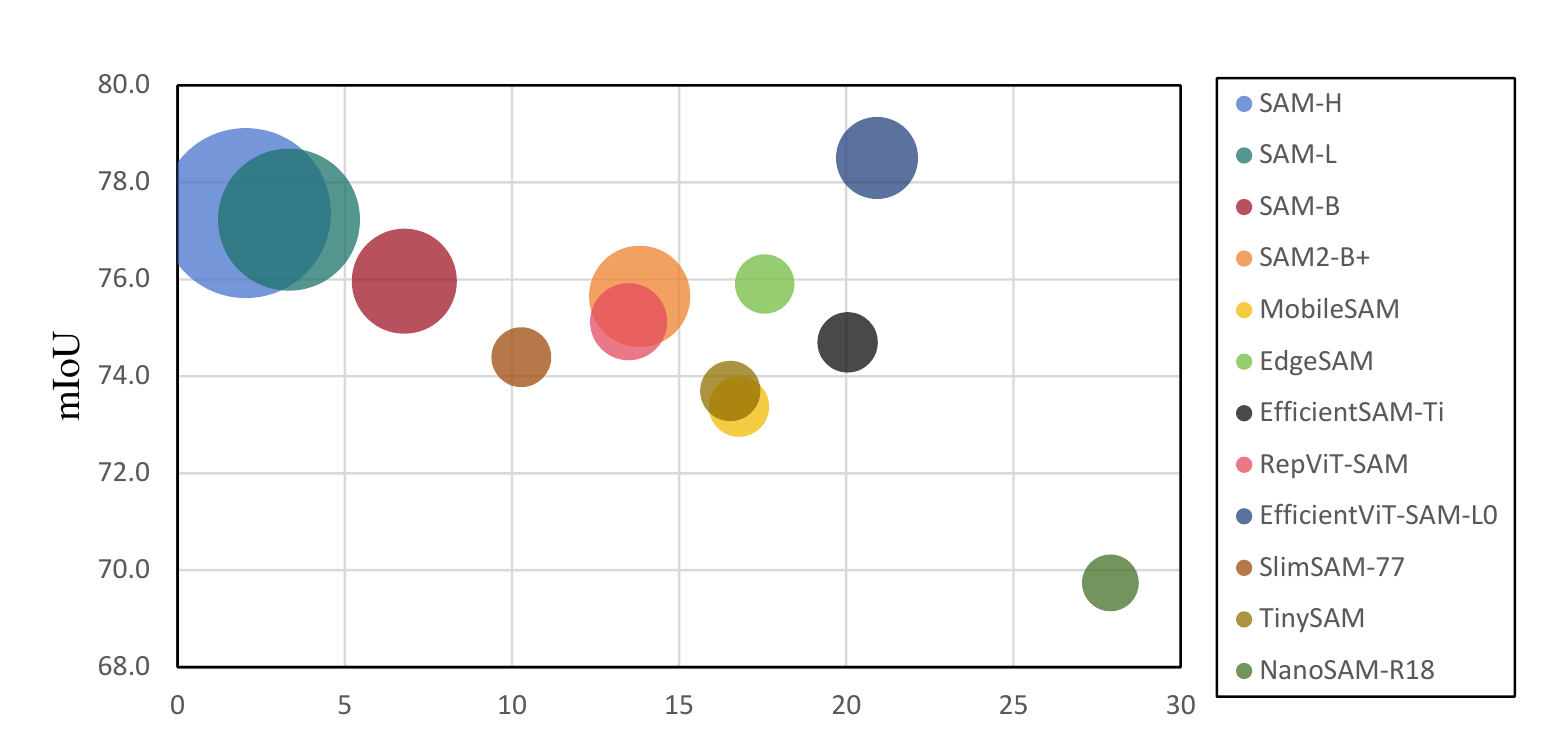}
    \caption{The trade-off between throughput and mIoU of SAM and its variants. The scale of circles responses to models' size.}
    \label{fig: throughput-miou}
\end{figure}

Based on the results in Section~\ref{sec: 4.2} and Section~\ref{sec: 4.3}, we further make a throughput-mIoU scatter to observe variants' efficiency-accuracy trade-off. Specifically, we select the throughput and mIoU that are evaluated on COCO dataset with ground truth bounding boxes as prompts. The results are illustrated in Fig. \ref{fig: throughput-miou}. 
For most of the variants, the improvement in latency accompanies with the degradation in accuracy.
For instance, NanoSAM at the bottom-right corner, demonstrates the highest throughput and lowest mIoU, while models around the top-left corner present poorer efficiency and better accuracy.
Besides, it is also worth noting that, EfficientViT-SAM-L0 on the top-right exhibits both the highest mIoU and relatively high throughput, achieving best efficiency-accuracy trade-off among these variants.

\section{Conclusion}
\label{conclusion}
In this survey, we have primarily discussed and evaluated the prominent works that focus on methods to efficiently segment anything and segment everything with reduced resource consumption and lower latency. For efficient SegAny tasks, most works adopt the approach of replacing either the image encoder or the entire architecture with a lightweight alternative, followed by training from scratch or through knowledge distillation. Other works aim to compress the original model by leveraging techniques such as quantization, pruning, or local optimization. For efficient SegEvery tasks, adopting an effective and efficient sampling strategy for prompt generation is essential.
After reviewing these methods in detail, we have also outlined five potential future research directions that may drive new trends in this area. Additionally, we have evaluated the efficiency, accuracy, and the corresponding trade-offs of these models in a consistent environment, providing a fair and valuable comparison. Our analysis shows that some variants have already outperformed the original SAM in specific scenarios, and we believe their success will inspire further exploration and innovation in this field.

\noindent\textbf{Data Availability.} The data supporting the experiments of this study are
openly available, as stated in Section \ref{sec: 4.1}.

\bibliography{sn-bibliography}

\end{document}